\documentclass[runningheads]{llncs}

\usepackage{eccv}

\usepackage{xspace}
\newcommand{\paper}{HKTex\xspace}

\usepackage{soul}
\setuldepth{foobar}

\usepackage{graphicx}
\usepackage{tikz}
\usepackage{multirow}
\usepackage{comment}

\usepackage{booktabs}
\usepackage{bm}
\usepackage{color}
\usepackage{xcolor}
\usepackage{fancyvrb}
\usepackage{algorithm,algpseudocode}
\algrenewcommand\algorithmicrequire{\textbf{Input:}}
\algrenewcommand\algorithmicensure{\textbf{Output:}}
\usepackage{mathrsfs}
\usepackage{makecell}
\usepackage{wrapfig}

\usepackage{eccvabbrv}
\usepackage[accsupp]{axessibility}  %
\usepackage{orcidlink}  %

\definecolor{cvprblue}{rgb}{0.21,0.49,0.74}
\usepackage[capitalize]{cleveref}

\usepackage{pifont}

\newcommand{\tm}{\mathrm{t}}
\renewcommand{\vec}[1]{\boldsymbol{#1}}
\newcommand{\mat}[1]{\mathbf{#1}}
\newcommand{\set}[1]{\mathcal{#1}}
\newcommand{\scal}[1]{\mathrm{#1}}

\newcommand{\Diag}{ {\mathbf{D}} }

\newcommand{\R}{\mathbb{R}}
\newcommand{\Nat}{\mathbb{N}}

\newcommand{\Id}{\mathbf{I}}

\renewcommand{\v}{\mathbf{v}}

\newcommand{\versor}[2][e]{
    \hat{\mathbf{#1}}
    \if\relax\detokenize{#2}\relax\else_{#2}\fi
}

\newcommand{\Loss}{\mathcal{L}} %

\newcommand{\hk}[1][\empty]{ 
    \ifx\empty#1 \vec{\kappa} \else 
    \ifx\relax#1\relax \vec{\kappa}_k \else 
    \vec{\kappa}_{#1} \fi \fi 
}
\newcommand{\Hks}{\mathcal{K}}
\newcommand{\anis}{\scal{\eta}}
\newcommand{\ang}{\scal{\theta}}
\newcommand{\thr}{\scal{\tau}}
\newcommand{\shr}{\scal{\varsigma}}
\newcommand{\rgb}{\mat{c}}

\newcommand{\Vertices}{\mathbf{X}}
\newcommand{\Edges}{\mathbf{E}}
\newcommand{\Faces}{\mathbf{F}}

\newcommand{\Mesh}{\Man}
\newcommand{\vt}[1][i]{\mathbf{x}_{#1}}

\newcommand{\face}[1][i]{\mathbf{f}_{#1}}
\newcommand{\faceindex}{\mathrm{f}}
\newcommand{\bary}{\mathbf{b}}

\newcommand{\ptx}[1][\empty]{ 
    \ifx\empty#1 \mathbf{p} \else 
    \ifx\relax#1\relax \mathbf{p}_j \else 
    \mathbf{p}_{#1} \fi \fi 
}  
\newcommand{\pt}[1][\empty]{ 
    \ifx\empty#1 \boldsymbol{p} \else 
    \ifx\relax#1\relax \boldsymbol{p}_j \else 
    \boldsymbol{p}_{#1} \fi \fi 
}
\newcommand{\src}{\pt^\star}
\newcommand{\Points}{\boldsymbol{P}}
\newcommand{\PointsPos}{\mat{P}}

\newcommand{\Lap}{\mathbf{L}}

\newcommand{\Lbo}{\Lap}
\newcommand{\Albo}{\Lap_{\anis, \ang}}
\newcommand{\Mass}{\mat{M}}
\newcommand{\Evecs}{\mat{\Phi}}
\newcommand{\Evals}{\mat{\Lambda}}
\newcommand{\evec}{\vec{\phi}}
\newcommand{\eval}{\lambda}
\newcommand{\efunc}{\varphi}
\newcommand{\aEvecs}{\Evecs_{\anis, \ang}}
\newcommand{\aEvals}{\Evals_{\anis, \ang}}
\newcommand{\aeval}{\eval^{\anis, \ang}}
\newcommand{\aefunc}{\efunc^{\anis, \ang}}

\newcommand{\Man}{\mathcal{M}}

\newcommand{\T}{\mathcal{T}}

\newcommand{\Exp}{\mathrm{{Exp}}}

\newcommand{\sota}{SotA}

\newcommand{\uv}[0]{$\mathrm{UV}$}

\newcommand{\mitsuba}{\mathtt{Mitsuba 3}}
\newcommand{\pytorch}{\mathtt{PyTorch}}
\newcommand{\drjit}{\mathtt{Dr.Jit}}

\newcommand{\faiss}{\mathtt{FAISS}}
\newcommand{\optuna}{\mathtt{Optuna}}

\crefname{equation}{eq.}{eq.}
\Crefname{equation}{Eq.}{Eq.}
\crefname{theorem}{thm.}{thms.}
\Crefname{Theorem}{Thm.}{Thms.}
\crefname{conjecture}{conj.}{conjs.}
\Crefname{Conjecture}{Conj.}{Conjs.}
\crefname{proposition}{prop.}{props.}
\Crefname{proposition}{Prop.}{Props.}
\crefname{definition}{dfn.}{dfn.}
\Crefname{definition}{Dfn.}{Dfn.}
\crefname{remark}{remark}{remark}
\Crefname{Remark}{Remark}{Remark}
\Crefname{algorithm}{Alg.}{Alg.}

\crefname{section}{Sec.}{Secs.}
\Crefname{section}{Sec.}{Secs.}
\crefname{equation}{Eq.}{Eqs.}
\Crefname{equation}{Eq.}{Eqs.}
\crefname{figure}{Fig.}{Figs.}
\Crefname{figure}{Fig.}{Figs.}
\crefname{table}{Tab.}{Tabs.}
\Crefname{table}{Tab.}{Tabs.}
\crefname{thm}{Thm.}{Thms.}
\Crefname{thm}{Thm.}{Thms.}
\crefname{conj}{Conj.}{Conjs.}
\Crefname{conj}{Conj.}{Conjs.}
\crefname{dfn}{Dfn.}{Dfns.}
\crefname{dfn}{Dfn.}{Dfns.}
\crefname{remark}{remark}{remarks}
\Crefname{Remark}{Remark}{Remarks}
\crefname{prop}{Prop.}{Prop.}
\Crefname{prop}{Prop.}{Prop.}
\Crefname{algorithm}{Alg.}{Alg.}
\crefname{appendix}{App.}{Apps.}
\Crefname{appendix}{App.}{Apps.}
\crefname{appsec}{appendix}{appendices}
\Crefname{appsec}{Appendix}{Appendices}

\renewcommand{\paragraph}[1]{{\vspace{1mm}\noindent \bf #1}}

\newlength{\savedintextsep}
\newlength{\savedcolumnsep}

\newcommand{\StoreParams}{
    \setlength{\savedintextsep}{\intextsep}
    \setlength{\savedcolumnsep}{\columnsep}
}

\newcommand{\StartTightFigureParagraph}{
    \StoreParams
    \setlength{\intextsep}{5pt}  
    \setlength{\columnsep}{7pt}
}

\newcommand{\EndTightFigureParagraph}{
    \setlength{\intextsep}{\savedintextsep}
    \setlength{\columnsep}{\savedcolumnsep}
}

\begin{document}
    \title{Heat Kernel Textures \\ the Geodesic Gaussians That Do Not Splat} 

\titlerunning{Heat Kernel Textures}
\newcommand{\samethanks}[1][\value{footnote}]{\footnotemark[#1] }
\author{Simone Foti\thanks{Equal contribution} \orcidlink{0000-0003-3207-1965} \and
Caner Korkmaz\samethanks \orcidlink{0000-0003-4066-7822} \and
Stefanos Zafeiriou \orcidlink{0000-0002-5222-1740} \and
Tolga Birdal \orcidlink{0000-0001-7915-7964}
}

\authorrunning{Foti et al.}

\institute{Imperial College London\\
}

    \maketitle
    
    \begin{figure}[h]
    \centering
    \vspace{-6mm}
    \includegraphics[width=\textwidth]{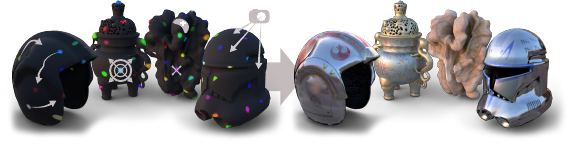}
    \vspace{-6mm}
    \caption{\paper: an intrinsic, \uv-free texture representation using heat kernels. \textit{Left:} A Riemannian framework constrains kernel movement to the surface, followed by manifold-aware controllers that manage kernel shape and density. Evaluation occurs within a differentiable ray-tracing pipeline. \textit{Right:} Final physically based renders.   \vspace{-5mm}
    }
    \label{fig:teaser}
\end{figure}

\begin{abstract}
    \vspace{-8pt}
    3D Gaussian Splatting has recently revolutionised novel view synthesis as well as many other 3D vision methods and applications. Drawing inspiration from this representation, we now rethink textures to overcome the main issues of UV mapping while considerably lowering their memory footprint. 
    Heat Kernel Textures (\paper) eliminate UV unwrapping as well as their persistent issues of wasted UV space, seams, distortions, vertex-duplication, and varying resolution.
    Grounded in discrete Riemannian geometry and intrinsically defined on any manifold surface discretised as a triangular mesh, \paper uses anisotropic heat kernels as geodesic equivalents to Gaussians. Like our kernels, also the optimisation of their position and the adaptive densification strategies were redefined to operate on the surface of the object to be textureised. Our novel representation is also fully integrated with a physically based renderer and can be optimised either from existing textures or multi-view images. Our project page and code are available at \href{http://circle-group.github.io/research/HeatKernelTextures}{circle-group.github.io/research/HeatKernelTextures}.
\end{abstract}

    \section{Introduction}
    \label{sec:intro}

    Rendering and representation techniques have evolved exponentially over the past decades, but the representation of objects' appearance is still mostly anchored in the traditional and fundamentally flawed \uv-mapping. With this representation, points on the surface of a mesh are mapped onto image planes where appearance properties are stored. Not only does this cause visual artifacts like seams, varying resolutions, and distortions, but it also incurs additional storage costs for saving the \uv-coordinates associated with every vertex of a mesh and for duplicating vertices on seam lines. Empty \uv-mapped regions further exacerbate memory waste. To overcome these limitations, we advocate for a new representation intrinsically defined on the surface of the objects and capable of lowering the memory footprint, with the potential to benefit games, virtual and augmented reality, and visual effects. 
    
    Multiple works have attempted to mitigate \uv-related issues, most notably through implicit neural representations~\cite{oechsle2019texture, koestler2022intrinsic} or coloured point clouds~\cite{yuksel2010mesh, yuksel2017meshcolourtextures, foti2024uv}. While promising, these approaches respectively incur considerable memory footprints, present topology-dependent structures, or require a substantial number of primitives. More recently, 3D Gaussian Splatting (3DGS)~\cite{kerbl20233d} has revolutionised novel view synthesis through a representation that tightly entangles geometry and appearance. However, standard 3DGS operates entirely in ambient Euclidean space. When applied to surface texturing, this assumption becomes a fundamental flaw: Euclidean optimisation naturally detaches Gaussians from the geometry, and the rigid 3D Gaussians themselves are incapable of bending to conform to the underlying manifold. While recent methods~\cite{guedon2024sugar, choi2024meshgs} attempt to align Gaussians to the surface through the introduction of additional regularisation losses, the fundamental disconnect remains.
    
    In contrast, we introduce a truly intrinsic, geodesic representation where the texturing primitives natively conform to the existing geometry of a mesh. Instead of operating in ambient 3D space, we replace Euclidean Gaussians with anisotropic heat kernels, defining all operations directly on the manifold surface. Specifically, kernel positions are updated through Riemannian geodesic optimisation, and our adaptive density controls also respect the manifold using the exponential map during kernel splits. To continuously and efficiently evaluate kernels across the surface, we devise new techniques to parametrise kernel anisotropies while simultaneously bridging continuous and discrete heat diffusion formulations. Our formulation enables seamless integration into physically based rendering pipelines where our textures can be differentiably ray-traced. It is important to note that our heat kernels are not splatted, but rather dynamically evaluated at ray intersections occurring during rendering. Without loss of generality, we focus this study on albedo texture properties, thus modelling how light reflects. We prove that Heat Kernel Textures (\paper) can compress existing UV textures and also be used for inverse rendering from 2D photos, thus enabling distortion- and seam-free texture acquisition by construction. 

    To summarise, our main contributions are: 
    \begin{enumerate}
        \item A \textbf{novel intrinsic texture representation} grounded in discrete Riemannian geometry that efficiently represents textures without \uv-mapping.
        \item A \textbf{Riemannian optimisation framework} that strictly constrains kernel movements and momentum to the surface of the mesh. 
        \item \textbf{Manifold-aware density controllers} specifically designed to operate alongside physically based rendering and our continuous texture evaluation.
        \item The \textbf{integration of \paper into a differentiable PBR} pipeline, enabling distortion-free texture optimisation from multi-view images. 
    \end{enumerate}

    \section{Related Work}
    \label{sec:related}

    \paragraph{Neural Texture Representations.}
        {\setlength{\emergencystretch}{0.2em} To circumvent the persistent artifacts of traditional \uv-mapping, substantial effort has been directed towards coordinate-based neural networks to represent surface appearance. Early implicit approaches evaluated Multilayer Perceptrons (MLPs) directly on 3D Euclidean coordinates~\cite{oechsle2019texture}. To improve spatial expressiveness, subsequent methods introduced hierarchical or multi-resolution per-vertex feature encodings that are interpolated and fed into MLPs~\cite{mahajan2024meshfeat}. Other hybrid approaches, such as NeuTex~\cite{xiang2021neutex}, attempted to map 3D points to an unfolded 2D space before neural evaluation. While these methods reduce reliance on manual \uv-unwrapping, they remain fundamentally anchored in ambient 3D space or intermediate 2D mappings. 
        Closely related to our mathematical foundation are Intrinsic Neural Fields~\cite{koestler2022intrinsic}, which bypass Euclidean coordinates by barycentrically interpolating the eigenvectors of the Laplace-Beltrami Operator (LBO) to condition an MLP. While mathematically elegant, MLP-based evaluations remain computationally heavy and lack the local, adaptive expressivity of explicit primitives. \par}
    
    \paragraph{Gaussian Splatting and Explicit Surfaces.}
        Point-based graphics and surface splatting have long been utilized for rendering~\cite{zwicker2001surface}, but they have seen a massive resurgence following the introduction of 3D Gaussian Splatting (3DGS)~\cite{kerbl20233d}. The unprecedented speed and visual fidelity of 3DGS have inspired attempts to adapt Gaussians to 2D domains, such as replacing image pixels with explicit 2D Gaussians~\cite{zhang2024gaussianimage, zhang2025imagegs, zhu2025large}. 
        To apply this expressive power to 3D meshes, recent works have attempted to bridge GS with texture mapping. However, these methods typically regress to utilizing \uv-parameterizations, either by storing Gaussians within the \uv-images of a sphere~\cite{rai2025uvgs} or by splatting within unfolded texture spaces~\cite{xu2024texturegs}. Other advanced hybrid representations, such as Neural Shell Textures~\cite{zhang2025neuralshelltexture}, use 2D Gaussians to represent local geometry while storing colors in multi-resolution hash grids~\cite{muller2022instant}. Crucially, all of these GS-based approaches either rely on explicit \uv-maps or operate in ambient Euclidean space, meaning their primitives cannot intrinsically bend or conform to general topological manifolds. 
    
    \paragraph{Integration with Physically Based Rendering.}
        While 3DGS was designed for rapid rasterisation, its integration into ray-traced physically based rendering (PBR) is an emerging frontier. Recent works have incorporated Gaussians into PBR frameworks for phenomena like particle rendering~\cite{condor2025don}, and extended them for complex inverse rendering and relighting tasks~\cite{gao2024relightable, ye2025geosplatting}. However, these approaches still rely on volumetric clouds of Euclidean primitives. Our work aligns with this trajectory but fundamentally shifts the paradigm: rather than rasterising ambient Gaussians, we natively integrate intrinsic, manifold-bound heat kernels as a continuous, differentiable texture evaluated directly at ray intersections within a PBR engine.

    \section{Background \& Notation}
    \label{sec:background}

    Let $\Mesh = \{\Vertices, \Faces\}$ be a manifold mesh with $\Vertices \in \R^{N \times 3} = [\vt[1] \dots \vt[N] ]^\top$ representing the positions of the $N$ vertices used to discretise the continuous surface of an object, and $\Faces \in \Nat^{F \times 3}$ the face matrix indexing how vertices are connected into faces. We also group edges in a matrix $\Edges \in \Nat^{E \times 2}$. 
    Points on the surface of $\Mesh$ are defined by the pair $\pt = (\faceindex, \bary)$, where $\faceindex \in [1, F]$ index the face $\face[\faceindex] \in \Faces$ on which a point lies, and $\bary = [\scal{b}_u, \scal{b}_v, \scal{b}_w]^T$ are its barycentric coordinates on $\face[\faceindex]$. 
    The coordinates of $\pt$ are barycentrically interpolated from the vertex coordinates $\vt[{\faceindex, 0}], \vt[{\faceindex, 1}], \vt[{\faceindex,2}]$ of $\face[\faceindex]$ as: $\ptx = b^{\pt} (\Vertices) =  \scal{b}_u \vt[{\faceindex, 0}] + \scal{b}_v \vt[{\faceindex, 1}] + \scal{b}_w \vt[{\faceindex, 2}]$.
    Similarly, barycentric interpolation can be used to interpolate any other attribute $\mat{A} \in \R^{N \times A}$ defined on the vertices of the mesh. 
    The simultaneous barycentric interpolation over multiple points $\Points = \{ \pt[] \}_{j=1}^J$, is hereafter indicated as:  $ b^{\Points} (\mat{A}) = \mat{A}^{\Points} \in \R^{P \times A}$. Therefore, $ b^{\Points} (\Vertices) = \Vertices^{\Points}$, which we also represent as $\PointsPos$ for simplicity.

    \StartTightFigureParagraph
    \paragraph{Laplace Beltrami Operators.} 
        We now briefly introduce the Laplace Beltrami operator (LBO) in both its isotropic and anisotropic~\cite{andreux2014anisotropic, melzi2016shape, wardetzky2007discrete} versions. We generally define the Laplacian as $\Lap(\anis, \ang) = -\Mass \mat{W}(\anis, \ang)$, where $\Mass$ is the diagonal \emph{mass matrix}, whose diagonal entries are proportional to the total area of the faces surrounding each vertex~\cite{sharp2020laplacian, foti2024uv}, and $\mat{W}(\anis, \ang)$ is the \emph{stiffness matrix} composed of weights:
        \begin{equation}
            \label{eq:stiff_lbo}
            \scal{w}_{ij} \! = \!
            \begin{cases}
              -\frac{1}{2} \Big[ \! \frac{\langle \versor{kj}, \versor{ki}\rangle_{\mat{H}_{\anis, \ang}}}{\sin (\alpha_{ij})} \! + \! \frac{\langle \versor{hj}, \versor{hi}\rangle_{\mat{H}_{\anis, \ang}}}{\sin (\beta_{ij})} \! \Big] & (i,j) \! \in \! \Edges, \\
              - \sum_{k \neq i} \scal{w}_{ik} & i\!=\!j,\\
              0 & \text{else}.
            \end{cases}
        \end{equation}
        The inner products between the edge versors are weighted by a \emph{shear matrix} $\mat{H}_{\anis, \ang} = \mat{R}_\ang \mat{U}_{ijk}\Diag_\anis \mat{U}_{ijk}^T \mat{R}_\ang^\top$.
        $\mat{U}_{ijk} = [\versor[u]{}, \versor[v]{}, \versor[n]{}]$ is the matrix containing a local reference frame defined on a face, with $\versor[n]{}$ being the normal to the surface and $\versor[u]{}, \versor[v]{}$ being the orthogonal components defined on the face ---a common choice involves using the principal curvatures and setting $\versor[u]{}=\versor[v]{M}$ and $\versor[v]{} = \versor[v]{m}$. 
        $\mat{R}_\ang$ is the rotation matrix rotating the reference frame, and $\Diag_\anis = \text{diag}( \frac{1}{1 + \anis}, 1, 1 )$ is a $3\times3$ 
        {\parfillskip0pt\par}
        \begin{wrapfigure}[7]{R}[0pt]{0.41\linewidth}
            \centering
            \vspace{-2pt}
            \includegraphics[width=\linewidth]{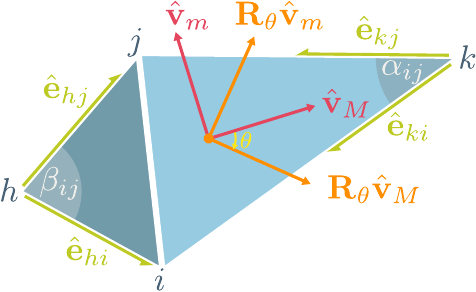}
        \end{wrapfigure}
        \noindent
        diagonal matrix controlling the anisotropy. 
        When $\anis = \ang =0$ we have $\mat{H}_{\anis, \ang} = \Id$, 
        thus $\langle \versor{kj}, \versor{ki}\rangle_{\mat{H}_{\anis, \ang}} = \versor{kj}^\top \mat{H}_{\anis, \ang} \versor{ki} = \cos (\alpha_{ij})$ and $\langle \versor{kj}, \versor{ki}\rangle_{\mat{H}_{\anis, \ang}} = \cos (\beta_{ij})$. 
        This reduces \cref{eq:stiff_lbo} to the traditional isotropic LBO formulation, which we indicate with $\Lbo = \Lap(0, 0)$. For $\anis, \ang \neq 0$ we have the anisotropic LBO (ALBO), which we indicate as $\Albo = \Lap (\anis, \ang)$.  

        The generalised eigendecomposition of the LBO can be written as $\mat{W} \Evecs = \Mass \Evecs \Evals$, where $\Evecs \in \R^{N \times K} = [\evec_1 \dots \evec_K]^\top$ represents the first $K$ eigenvectors defined at the vertices of the mesh, and $\Evals = \text{diag}(\eval_1,\dots,\eval_K)$ is the diagonal matrix containing the eigenvalues. The eigendecomposition of the ALBO follows the same procedure to determine the $\anis$- and $\ang$-dependent $\aEvecs$ and $\aEvals$.
        \EndTightFigureParagraph

    \paragraph{Heat Diffusion and Kernels.}
        The heat diffusion process over a mesh is mathematically described as: $\Lap \; u(\vt, \tm) = \frac{\partial u(\vt, \tm)}{\partial t}$, 
        where the vertex-valued scalar function $u: \Mesh \times \R^+ \rightarrow \R$ represents how an initial heat distribution over $\Mesh$ evolves for times $\tm > 0$. Indicating with $\Vec{u}_\tm$ the vector obtained by evaluating $u(\vt, \tm)$ for all vertices, the solution of the heat equation has the form~\cite{sharp2022diffusionnet, foti2024uv, sun2009concise, chung2015unified}:
        \begin{equation}
            \label{eq:heat_kernel}
            \Vec{u}_\tm = e^{-t\mathbf{L}} \Vec{u}_0 \approx \mathbf{\Phi} e^{-t\mathbf{\Lambda}} \mathbf{\Phi}^T \mathbf{M} \Vec{u}_0 
        \end{equation}
        When the initial distribution is a delta function, $u(\vt, 0) = \delta(\vt) = \Vec{\delta}_{\vt}$, \cref{eq:heat_kernel} is referred to as \emph{heat kernel} and it represents how the heat generated by a single point source diffuses over the surface. In Euclidean space, it has the explicit expression of a Gaussian~\cite{sharp2019vector}.
        Note that while it is possible to simulate the diffusion by solving a sparse linear system, the spectral approach we adopt proved to be one order of magnitude faster and to scale to high-resolution meshes~\cite{sharp2022diffusionnet}.

\section{Heat Kernel Textures}
    \label{sec:hkt}
    \StartTightFigureParagraph
    We now detail our \paper model, which uses arbitrarily positioned heat kernels to intrinsically represent textures directly on the surface of manifold meshes.
        
    \paragraph{Kernel parameters.} 
        We define our \paper as a collection of $\scal{S}$ optimisable heat kernels, $\Hks = \{\hk[s]\}_{s=1}^\scal{S}$. Each kernel is parametrised as $\hk = \big(\src, \ang, \anis, \thr, \shr, \rgb  \big)$,
        {\parfillskip0pt\par}
        \begin{wrapfigure}[7]{L}[0pt]{0.41\linewidth}
            \centering
            \vspace{-10pt}
            \includegraphics[width=\linewidth]{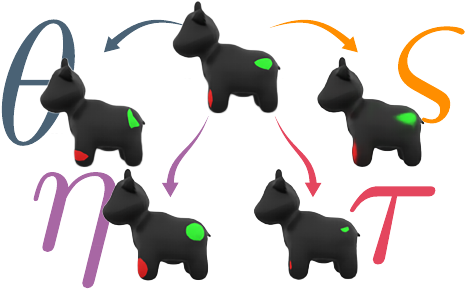}
        \end{wrapfigure}
        \noindent
        where $\src$ is the position of the heat source, $\ang$ and $\anis$ are the angle and anisotropy controlling the diffusion process, $\thr$ is the threshold controlling the scale of the kernel, $\shr$ is the parameter controlling the sharpness of its border, and $\rgb$ is the RGB colour. The parameters of all $\Hks$ are jointly optimised.
    \EndTightFigureParagraph

    \paragraph{Normalised heat kernels.} 
        In \cref{sec:background} we have introduced the discrete formulation of heat kernels and diffusion. While this is the gold standard on 3D meshes, we seek to break free from the mesh discretisation and represent textures beyond vertices. We thus leverage the continuous manifold formulation of heat kernels and define how to transform discrete quantities into continuous ones. Anisotropically heat diffusing a Dirac delta function centred in $\src$ for a time $t$, we have the truncated continuous anisotropic heat kernel:
        \begin{equation}
            \label{eq:cont_heat_kernel}
            h_t\big(\pt, \src \! \mid \ang, \anis\big) = \sum_{k=1}^K \; e^{-t\aeval_k } \; \aefunc_k\!(\pt) \; \aefunc_k\!(\src) \; m^\star,
        \end{equation}
        where $\pt$ is any arbitrary point on the surface of $\Mesh$, $\aeval_k$ is the $k$-th eigenvalue with its corresponding eigenfunction $\aefunc_k$ evaluated at $\pt$ or $\src$, and $m^\star$ is the mass at the source. 

        Since the physical heat diffusion process conserves energy, as heat spreads, the source temperature decreases. Therefore, the size and shape of kernels would affect their overall temperature. To stabilise optimisation and make filtering predictable, we guarantee a maximum temperature of $1.0$ at the kernel centre by normalising with respect to the post-diffusion source temperature:  
        \begin{equation}
            \bar{h}_t\big(\pt \! \mid \src \!, \ang, \anis\big) = \frac{h_t\big(\pt, \src \! \mid \ang, \anis\big)}{h_t\big(\src \!, \src \! \mid \ang, \anis\big)}.
        \end{equation}

    \paragraph{Biharmonic distance weighting.}
        Heat kernels have a global support across the entire surface of the object. While they should decay with distance from the heat source, since we adopt a spectral truncation to the $K$ most significant frequencies, the kernels suffer from a Gibbs-like phenomenon \cite{patane2016star, patane2023spectral} producing \textit{spectral ringing} artifacts in distant regions (see supplementary materials). To suppress these unwanted artifacts and enforce local support, we introduce a Biharmonic distance Gaussian weighting that dampens the contributions of distant kernels. We compute the Biharmonic distance~\cite{lipman2010biharmonic} using isotropic eigenproperties:
        $d_\text{BH}(\pt, \src)^2 = \sum_{k=1}^K \frac{(\efunc_k^\text{iso}(\pt) - \efunc_k^\text{iso}(\src))^2}{(\eval_k^\text{iso})^2}$. 
        Then, we weight our normalised kernels:
        \begin{equation}
            \label{eq:bhd_weighting}
            \tilde{h}_t(\pt \! \mid \src \!, \ang, \anis) = %
            \bar{h}_t(\pt \! \mid \src \!, \ang, \anis) \cdot \exp \! \left( \! - \frac{d_\text{BH}(\pt, \src)^2}{2\sigma_\text{BH}^2} \right).
        \end{equation}

    \paragraph{Kernel filtering.}
        $\tilde{h}_t(\pt \!\mid \src \!, \ang, \anis)$ produces a Gaussian-like geodesic kernel, with values in $[0,1]$, that gradually fades from the source location. Instead of optimising the diffusion time to scale the kernel, we fix $t$ and modulate kernel scale ($\thr$) and sharpness ($\shr$) via a rescaled soft-step filtering function:  
        \begin{equation}
            \label{eq:hk_filtering}
            \alpha ( \pt\! \mid \src \!, \ang, \anis, \thr, \shr ) = \frac{\xi\big( \tilde{h}_t(\pt \! \mid \src \!, \ang, \anis) \big) - \xi(0)}{\xi(1) - \xi(0)}, \quad \text{with} \quad \xi(x) = \sigma \big( \shr \cdot ( x - \thr ) \big),
        \end{equation}
        where $\sigma(\cdot)$ is a sigmoid function.
        This filtering preserves the $[0,1]$ value range and provides more control over the kernel appearance, which can now be easily scaled down to point-sized dimensions while controlling the rate of signal decay. A filtered kernel can now either preserve its characteristic fading appearance or transform into a crisp geodesic ellipsoid defined on the surface of the mesh.

    \paragraph{Colour formation.}
        For a given kernel $\hk[s] \in \Hks$, we denote its evaluated spatial footprint as $\alpha_s(\pt) \equiv \alpha(\pt \mid \src_s, \ang_s, \anis_s, \thr_s, \shr_s)$. To determine the final texture colour $C(\pt)$ at a surface point $\pt$, we need to aggregate the contributions of all kernels. Inspired by \cite{zhang2025imagegs}, we select the $\scal{K}_c$ top-contributing kernels to $\pt$. In our formulation, this can be achieved by dynamically identifying the kernels with the highest $\alpha_s(\pt)$. The local texture is then computed via an alpha-weighted partition of unity over these selected kernels ($\Hks(\pt) \subset \Hks$), which is added as a residual to an optimisable global base colour $\rgb_\text{base}$:
        \begin{equation}
            \label{eq:color_formation}
            C(\pt) = \rgb_\text{base} + \frac{\sum_{s \in \Hks(\pt)} \alpha_s(\pt) \; \rgb_s}{\sum_{s \in \Hks(\pt)} \alpha_s(\pt)},
        \end{equation}
        This colour formation paradigm guarantees a controlled blending between locally overlapping kernels, while empty regions seamlessly default to the underlying base colour.

        \StartTightFigureParagraph
    \subsection{Parametric Interpolation of the Spectral Bases}
        \label{sec:precomp_interp_eigen}
        \paragraph{Interpolation of ALBO eigenproperties.} 
            {\setlength{\emergencystretch}{1.2em} Evaluating the continuous anisotropic heat kernel in \cref{eq:cont_heat_kernel} requires the spectral properties associated with a kernel's specific shape parameters $(\anis, \ang)$. Since constructing and eigendecomposing a unique ALBO for every kernel during an optimisation pipeline is computationally prohibitive, we devise a simplified and easily differentiable alternative.
            We thus discretise the parameter space of angles and anisotropies into a half-polar-like grid, where anisotropies are log-scaled to make the space more perceptually uniform and prevent high anisotropies from dominating.
            We associate each grid node $(i,j)$, located at discrete angle $\ang_i$ and log-scaled anisotropy $s_j = \ln(1 + \eta_j)$, with its precomputed $\Evecs_{i,j}^\ddag$ and $\Evals_{i,j}^\ddag$. To evaluate a continuous 
            {\parfillskip0pt\par}
            \begin{wrapfigure}[9]{R}[0pt]{0.38\linewidth}
                \centering
                \vspace{-5pt}
                \includegraphics[width=\linewidth]{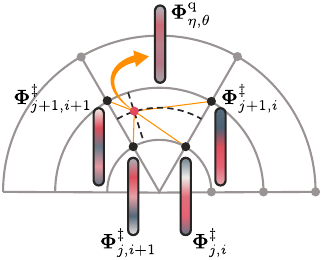}
            \end{wrapfigure}
            \noindent
            kernel query with parameters $(\anis, \ang)$, we first project the queried anisotropy into the same logarithmic space, $s = \ln(1 + \anis)$. We then perform a binary search to identify the bounding grid cell indices $i$ and $j$ such that $\ang_i \le \ang < \ang_{i+1}$ and $s_j \le s < s_{j+1}$. Within this cell, we compute the normalized local coordinates: $\Delta\ang = \frac{\ang - \ang_i}{\ang_{i+1} - \ang_i}$ and $\Delta s = \frac{s - s_j}{s_{j+1} - s_j}$.        
            The queried eigenproperties are then efficiently approximated via bilinear interpolation over the four cell corners $\set{C} = \{ (a,b) \! \mid a \in \{j, j+1\}, b \in \{i, i+1\} \}$:
            \begin{equation}
                \label{eq:eigen_bilinterp}
                \aEvecs^\text{q} = \sum_{(a,b) \in \set{C}} w_{a,b} \Evecs_{a,b}^\ddag \qquad \text{and} \qquad \aEvals^\text{q} = \sum_{(a,b) \in \set{C}} w_{a,b} \Evals_{a,b}^\ddag,
            \end{equation}
            where the interpolation weights $w_{a,b}$ are derived from the fractional distances $\Delta\ang$ and $\Delta s$ (i.e., $w_{i,j} = (1-\Delta\ang)(1-\Delta s)$, $w_{i+1,j} = \Delta\ang(1-\Delta s)$, etc.). \par}
            \EndTightFigureParagraph

        \paragraph{Alignment of precomputed ALBO eigenproperties.}
            When eigendecomposing the precomputed ALBOs, we observe the same permutation, rotations, and sign ambiguities that are widely observed in shape matching problems (e.g.,~\cite{ovsjanikov2012functional}). Since this behaviour would hinder the correctness of the bilinear interpolation in \cref{eq:eigen_bilinterp}, we introduce an eigenproperties alignment procedure during the grid construction and thus ensure that all $\Evecs_{i,j}^\ddag$ and $\Evals_{i,j}^\ddag$ are consistent with the $\Evecs$ and $\Evals$ of the isotropic LBO.  
            First, we resolve the permutation switching issue by maximizing the mass-weighted cross-correlation between two sets of eigenvectors: $\Evecs_{\text{ref}}$ and $\Evecs_{i,j}^\ddag$. We compute the correlation matrix $\mat{C} = \Evecs_{\text{ref}}^\top \Mass \Evecs_{i,j}^\ddag$ and apply the Hungarian algorithm to find the optimal assignment matrix $\mat{\Pi}$, yielding the permuted eigenvectors $\Evecs_{\text{perm}} = \Evecs_{i,j}^\ddag \mat{\Pi}$. Next, we solve the Orthogonal Procrustes problem minimising the mass-weighted Frobenius distance $\| \Evecs_{\text{ref}} - \Evecs_{\text{perm}} \mat{R} \|_{\Mass}^2$ via the Singular Value Decomposition (SVD) of $\Evecs_{\text{perm}}^\top \Mass \Evecs_{\text{ref}}$. The resulting rotation determines $\Evecs_{\text{rot}} = \Evecs_{\text{perm}} \mat{R}$. Finally, we correct the sign flips by constructing a diagonal matrix $\mat{S}$ where $\mat{S}_{k,k} = \text{sgn}( (\Evecs_{\text{ref}}^\top \Mass \Evecs_{\text{rot}})_{k,k} )$. The fully aligned spectral basis stored at each grid node is thus obtained as $\Evecs_{\text{aligned}} = \Evecs_{i,j}^\ddag \mat{\Pi} \mat{R} \mat{S}$. Eigenvalues are simply reordered according to $\mat{\Pi}$. The initial reference eigenvectors are those of the LBO, then we use the last set aligned during the grid construction. Hereafter, we consider all $\Evecs_{i,j}^\ddag$ and $\Evals_{i,j}^\ddag$ to be implicitly aligned for ease of notation.

    \subsection{Spatial Evaluation of Spectral Properties and Masses}
        \label{sec:spatial_eigen_and_masses}

        \paragraph{Continuous spatial evaluation.}
            {\setlength{\emergencystretch}{0.2em} All precomputed spectral properties are strictly defined on the vertices of $\Mesh$. Limiting kernel placement and evaluation exclusively to these locations would bind our model to the surface discretisation. To overcome this limitation and evaluate any vertex-based spectral quantity $\mat{A}$ at an arbitrary continuous location $\pt$, we exploit the inherent smoothness of the spectral bases. Inspired by \cite{foti2024uv}, we approximate this continuous spatial evaluation via barycentric interpolation: $b^{\pt}(\mat{A})$. \par}

        \paragraph{Biharmonic KNN search.}
            Evaluating the heat equation for all $\scal{S}$ kernels and $\scal{P}$ target points using \cref{eq:cont_heat_kernel}, yields an intractable $\mathcal{O}(P \times S)$ complexity. Because our biharmonic distance weighting (\cref{eq:bhd_weighting}) forces kernels to decay to zero outside a local radius, we can safely prune distant sources. We define a spectral embedding at the vertices as $\mat{Z} = \Evecs^\text{iso} (\Evals^\text{iso})^{-1}$. In this space, Euclidean distance corresponds to the biharmonic distance $d_\text{BH}$ on the manifold. Using our continuous evaluation strategy, we compute these embeddings for kernel sources and target points as $\mathbf{z}(\pt) \approx b^{\pt}(\mat{Z})$ and $\mathbf{z}(\src) \approx b^{\src}(\mat{Z})$. We then construct a K-Nearest Neighbours (KNN) graph in this embedded space to dynamically restrict the evaluation of each point $\pt$ to only its $\scal{K}_s$ nearest sources $\mathcal{N}(\pt)$.
        
        \paragraph{Local anisotropic evaluation.}
            Once the local subset of sources $\mathcal{N}(\pt)$ is identified, we must evaluate the continuous anisotropic eigenfunctions specifically required by those sources. Rather than computing the basis for all kernels globally, we retrieve the queried discrete eigenvectors $\aEvecs^\text{q}$ corresponding to the specific shape parameters $(\ang_s, \anis_s)$ of only the $\scal{K}_s$ neighbouring sources. Applying our continuous evaluation principle, we approximate these source-specific eigenfunctions at             both the source location $\src$ and the target location $\pt$. Specifically, we compute $[\aefunc_1(\src), \dots, \aefunc_K(\src)] \approx b^{\src}(\aEvecs^\text{q})$ and $[\aefunc_1(\pt), \dots, \aefunc_K(\pt)] \approx b^{\pt}(\aEvecs^\text{q})$. This dual evaluation allows us to construct the heat kernel between $\src$ and $\pt$ (\cref{eq:cont_heat_kernel}), while keeping the computational complexity strictly bounded to $\mathcal{O}(P \times \scal{K}_s)$.
    
        \paragraph{Mass definition for arbitrary points on $\Mesh$.}
            Finally, evaluating the continuous heat kernel (\cref{eq:cont_heat_kernel}) requires the mass $m^\star$ at the continuous source point $\src$. Since the discrete mass matrix ($\Mass$) is proportional to vertex areas and thus local sampling density, we cannot simply use barycentric interpolations like we did for the eigenvectors.%
            We thus formulate a Kernel Density Estimation (KDE) approach to correctly estimate the mass of every point based on the density of surrounding points (see supplementary materials). While this approach approximates the real mass, we empirically find that assigning a uniform unit mass to all sources ($m^\star = 1$) is significantly faster and does not deteriorate performance. %
            Because the pointwise evaluation depends solely on the source mass as a linear scalar (\cref{eq:cont_heat_kernel}), during optimisation the other kernel properties can learn to compensate for the missing local integration weights.

    \subsection{Optimisation \& Density Control}
    
        \paragraph{Geodesic position optimisation.}
            Standard Euclidean optimisation would apply gradient updates directly to the 3D coordinates, inevitably pulling the heat kernel sources off the mesh surface. Since sources $\src$ are intrinsically defined on $\Mesh$, we must restrict movements exclusively to the surface. We thus formulate a Riemannian gradient descent optimiser: each gradient $\nabla_{\src} \mathcal{L}$ is projected to the local tangent plane $\T_{\src}\Mesh$ and scaled to define an initial velocity $\v^\star$. We then compute the exponential map $\Exp(\src, \v^\star)$, tracing the geodesic defined by this vector to determine the updated kernel position. To accelerate convergence with momentum, we parallel transport the accumulated momentum to the new kernel position. Both the exponential map and parallel transport are computed via the straightest geodesic algorithm~\cite{polthier2006straightestgeod}, utilising the optimisation-friendly and CUDA-parallelised implementation introduced by~\cite{verninas2026disgeod}.
    
        \paragraph{Kernel pruning strategy.}
            Like most Gaussian Splatting techniques, we adaptively control the density of our heat kernels throughout the optimisation process. In order to decrease computational overhead and storage costs, we aim to delete kernels that contribute little to the final texture. During an accumulation window $\Delta_\text{prune}$, we record two metrics for every kernel $\hk_s \in \Hks$: a \textit{hit count} $\nu(\hk_s)$ indicating how frequently it is selected in the local partition of unity $\Hks(\pt)$ (\cref{eq:color_formation}), and an \textit{energy score} $\omega(\hk_s)$ aggregating its filtered weights $\alpha_s(\pt)$. Periodically, we prune kernels that are either rarely hit ($\nu(\hk_s) < \gamma_\text{hit} \cdot \max(\{\nu(\hk_i)\}_{i=1}^\scal{S})$) or energetically weak ($\omega(\hk_s) < \gamma_\text{energy}$).

        \paragraph{Kernel densification strategy.}
            To capture missing details, we dynamically densify kernels based on local reconstruction error. Over an accumulation window $\Delta_\text{error}$, we compute the spatial error $e(\pt) = \| \rgb(\pt) - \rgb_\text{GT}(\pt) \|_1$ and distribute it to contributing kernels weighted by their normalised partition of unity, yielding a per-kernel score $E(\hk_s)$. Kernels exceeding an error threshold $\gamma_\text{error}$ are densified according to their spatial extent ($1 - \tau_s$). Small kernels ($1 - \tau_s \le \gamma_\text{size}$) are \textit{cloned} in place, allowing the optimiser to naturally separate them. Large kernels ($1 - \tau_s > \gamma_\text{size}$) are \textit{split} into two smaller kernels. Crucially, to respect manifold constraints during splitting, we displace the new kernels along their principal axis using the exponential map provided by~\cite{verninas2026disgeod}. A maximum growth ratio $\gamma_\text{growth}$ caps additions per step to ensure stability.

    \subsection{Rendering \& Applications}
        \paragraph{Physically based rendering.}
            Because our \paper formulation is continuously evaluated across $\Mesh$, it seamlessly integrates into modern physically based rendering (PBR) pipelines. We implement \paper also as a differentiable texture within the $\mitsuba$ renderer~\cite{jakob2022mitsuba3}. During ray tracing, when a ray intersects the mesh at an arbitrary continuous point $\pt$, the renderer queries the corresponding local subset of kernels $\Hks(\pt)$ and computes $C(\pt)$ following \cref{eq:color_formation}. By wrapping our $\pytorch$-based evaluation within $\drjit$~\cite{jakob2022drjit}, we allow gradients to backpropagate directly from the rendered image pixels to our intrinsic kernel parameters.
    
        \paragraph{Optimise existing UV-textures.}
            When a ground-truth UV texture is available, we can directly optimise our kernels. At each optimisation step, we uniformly sample a dense set of points $\pt$ directly on the faces of $\Mesh$, and evaluate their colour with $C(\pt)$. Our \paper model is then updated by minimizing the photometric error with respect to $C_\text{GT}(\pt)$, the colour queried from the UV texture at the same location: $\mathcal{L} = \| C(\pt) - C_\text{GT}(\pt) \|_2^2$. 

        \paragraph{Optimise multi-view images.}
            Our \paper model can also be optimised from multi-view images via the differentiable PBR renderer. In this scenario, we suppose the target mesh to be given alongside a set of camera poses. At each training step, we sample $\scal{V}$ camera views, identify all the rays hitting the mesh across views, and select a random subset to provide uniform coverage around the object. Minimising the rendering error against the ground-truth pixels at the ray intersections drives the \paper optimisation.

    \section{Experiments}
    \label{sec:experiments}

    \subsection{Data}
        We conduct experiments on a curated subset of Objaverse~\cite{deitke2023objaverse}. Since the full dataset comprises scraped online assets, mesh quality is not guaranteed.
        We therefore filter for objects that possess LVIS category labels, valid textures, and at least one community like and view, while excluding point clouds and animated models.
        To maintain a reasonable computational overhead, we restrict the maximum vertex count to $N=60,000$. To ensure compatibility with our \paper representation, we also define a rigorous geometric filtering: we exclude meshes with multiple disconnected components, with non-manifold geometries, and with complex internal structures (i.e. models presenting more than $20$ intersections on any ray cast from a regular grid defined on the bounding box of the object). Finally, since our method is heavily reliant on the eigenfunctions of LBO, we reject models where such eigendecomposition fails. After filtering, we retain $\sim 3,000$ high-quality models. %
        Note that while having an eigendecomposable LBO is essential, our filtering criteria can be relaxed at the cost of overall dataset quality or of separately processing parts of non-manifold and complex meshes.
        When fitting multi-view images, meshes are rendered with their albedos as the sole texture map.

    \subsection{Implementation details}

        Our \paper pipeline is built with $\pytorch$~\cite{paszke2019pytorch}, utilizing the GPU version of $\faiss$~\cite{johnson2019faiss} for the biharmonic distance KNN search and $\mitsuba$~\cite{jakob2022mitsuba3} with $\drjit$~\cite{jakob2022drjit} for differentiable ray tracing. Instead of building specialized CUDA kernels, we use $\pytorch$'s Triton code generation~\cite{ansel2024pytorch2}. For all meshes, we precompute a 64-dimensional eigendecomposition of the isotropic LBO for calculating biharmonic distances, and 256-dimensional eigendecompositions for all the ALBOs in our grid with the cross product of 7 angles and 7 anisotropies. The network parameters are initialized as in the following paragraph, and we use a projected gradient descent (PGD) algorithm to enforce the correct ranges. For all parameters except kernel locations, we use an Adam optimiser with a step-function learning rate scheduler for fitting UV-textures and multi-view images. To optimize kernel locations, we utilize a geodesic Riemannian-SGD algorithm on the GPU using the library from~\cite{verninas2026disgeod}. We run a short Bayesian hyperparameter tuning on 5 meshes excluded from the evaluation dataset for both setups using $\optuna$~\cite{akiba2019optuna} with the Tree-Structured Parzen Estimator~\cite{bergstra2011tpe}, and we share the exact hyperparameters in our supplementary. We use an importance sampling strategy, as detailed in our supplementary material, when fitting UV-textures. To render our output images and to perform differentiable ray tracing for inverse rendering, we utilize $\mitsuba$~\cite{jakob2022mitsuba3} with $\drjit$~\cite{jakob2022drjit}. We use a path-replay backpropagation (PRB) integrator with a ray depth of 3 for training, and differentiate through the ray tracing process to optimize our textures. During inference, we only build the KNN database and pre-compute the self-reference heat kernels once, reducing computation significantly. We use the tuned hyperparameters to set the pruning and densification hyperparameters for both settings.

        \paragraph{Initialisation.} We use a uniform sampling strategy based on mesh area to first oversample a large number of points on the mesh surface, and then use farthest-point sampling to obtain kernel locations and face indices. The kernel colours are sampled from a uniform distribution between -1 and 1, with the mean subtracted and the mean set to our global base colour. The angles and anisotropies are sampled from a uniform distribution between 0 and $\pi$ and 1 and 100, respectively. For sharpness and threshold initialization, we sample from the beta distributions $\mathcal{B}(2, 1)$ and $\mathcal{B}(10/7, 1)$, respectively. Then, the sharpnesses are scaled and shifted to the range 10-50, and the thresholds to 0.9-1.0 (0.7-1.0 for the multi-view rendering case). The parameters are then optimized, as detailed before and in the supplementary, using PGD.

    \subsection{Evaluation metrics}
        To quantitatively assess textures against the ground truth, we render five distinct views for each object and measure image metrics on them. In particular, we compute Mean Squared Errors (MSE) and Peak Signal-to-Noise Ratio (PSNR). To evaluate the preservation of structural details and contrast, we use the Structural Similarity Index (SSIM) and its multi-scale variant (MS-SSIM). Furthermore, we report the Learned Perceptual Image Patch Similarity (LPIPS) to measure perceptual quality. Finally, to validate our core claim of memory efficiency, we report the total Storage footprint in kilobytes (KB), strictly accounting for all intrinsic parameters, network weights, and any topological overheads (e.g., \uv-coordinates) required by the baselines. We store with the same formats and without method-specific quantizations. Identical {\sc zip} compression was already applied across all methods to provide a fair baseline for comparing intrinsic information density.
        \begin{figure}[t]
            \centering
            \includegraphics[width=1\linewidth]{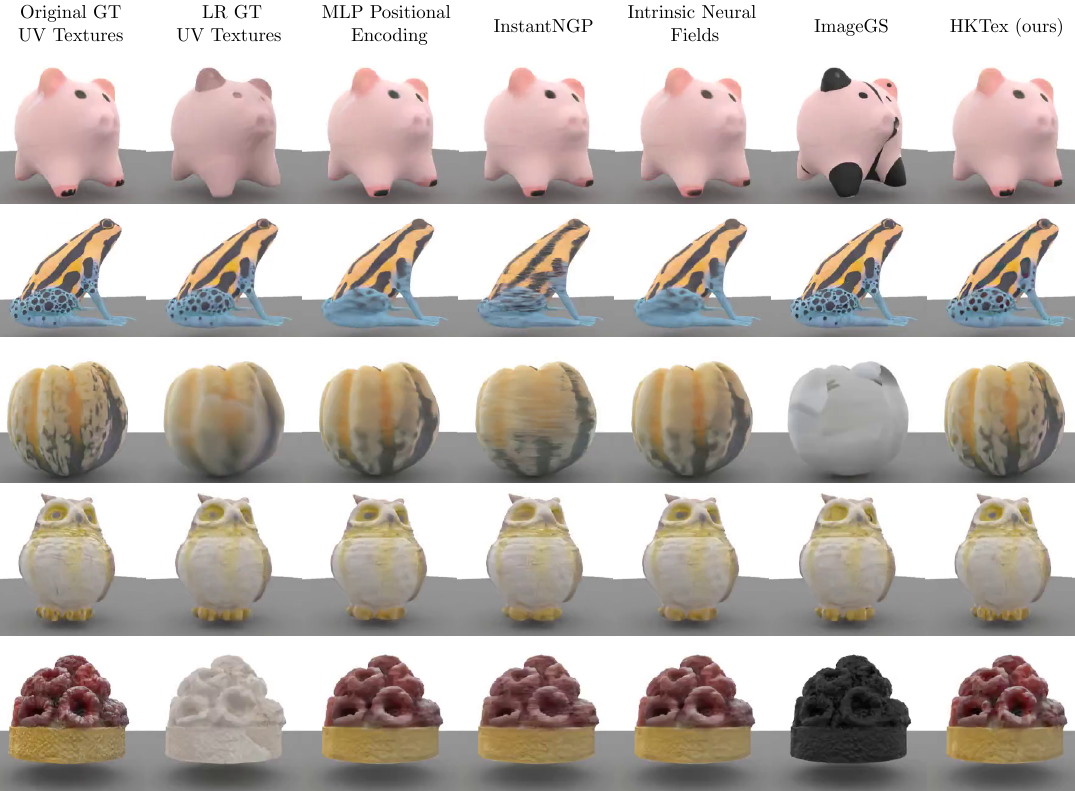}
            \caption{Qualitative comparison against \sota~ methods. With the exception of the GT \uv Textures, all other methods have similar or bigger storage size. For visual clarity, we render only the albedo textures under constant ambient illumination. %
            }
            \label{fig:qual results}
        \end{figure}

        \StartTightFigureParagraph
    \subsection{Fitting \uv-textures}
        We first demonstrate our method's ability to represent existing UV textures at a significantly reduced storage cost. We optimise our \paper against the ground truth albedo \uv-texture of $313$ meshes, arbitrarily selected from our curated dataset. We then quantitatively and qualitatively compare against leading state-of-the-art intrinsic texture representations and image-based baselines under strictly matched memory budgets. 
        
        For the neural baselines, we implement three multilayer perceptrons (MLPs) taking as input different positional encodings of the ray-intersection points. Specifically, we evaluate a standard MLP with traditional Fourier positional encoding~\cite{mildenhall2021nerf}, an adaptation of InstantNGP~\cite{muller2022instant} utilizing hash-grid encoding, and Intrinsic Neural Fields~\cite{koestler2022intrinsic}, which barycentrically interpolates the eigenvectors of the LBO. To ensure a fair comparison, we conduct a grid search over the network width, depth, and hash-grid size to exactly match the storage cost of our optimised \paper.
                
        Furthermore, we compare against two explicit 2D baselines: downsampled ground-truth images and ImageGS~\cite{zhang2025imagegs} (a 2D Gaussian Splatting approach). Crucially, both image-based methods fundamentally rely on \uv-mapping. Even when the \uv map is provided, the storage of per-vertex \uv-coordinates incurs a fixed memory overhead that must be accounted for in the total texture budget. Consequently, for meshes with high vertex counts, this overhead dominates the budget, severely restricting the memory available for the actual texture or 2D Gaussians. As shown in \cref{tab:uv-fitting}, an exact storage match is often impossible for these baselines; even when limiting the downsampled texture to a minimum resolution of $4\times4$ pixels and ImageGS to just $10$ Gaussians, the performance penalty is catastrophic for densely sampled meshes (\cref{fig:qual results}).

        \begin{wrapfigure}[5]{R}[0pt]{0.24\linewidth}
            \centering
            \vspace{-19pt}
            \includegraphics[width=\linewidth]{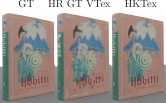}
            \vspace{-22pt}
            \caption{Comparison with vertex colours.}
            \label{fig:vs-vtex}
        \end{wrapfigure}
        Finally, we compare against vertex colours directly sampled from the GT \uv-textures at the native mesh resolution (GT VTex) as well as iteratively subdividing the mesh to match our storage budget (HR GT VTex).

        As reported in \cref{tab:uv-fitting}, \paper achieves the lowest storage footprint after GT VTex while maintaining competitive reconstruction quality across all metrics. Although LR \uv-textures achieve the highest PSNR due to directly storing image samples, our representation yields the second best perceptual quality (LPIPS) and the best SSIM, MS-SSIM, and MSE, indicating improved preservation of structural details despite using fewer parameters. Neural baselines and HR GT VTex require significantly larger storage budgets to achieve almost comparable accuracy, highlighting the efficiency of our geodesic heat kernel representations.

        The qualitative comparisons in \cref{fig:qual results} further highlight the behaviour of the different representations. Neural field baselines tend to oversmooth fine texture details, particularly in regions with sharp colour transitions or high-frequency patterns (e.g., the frog and raspberry examples). ImageGS occasionally produces noticeable artifacts due to the limited number of splats available under the storage constraints, sometimes failing to capture the correct colour distribution. Vertex colours, even when subdivided (\cref{fig:vs-vtex}), fail to correctly represent textures because the vertex positioning is determined by geometry rather than texture content. In contrast, \paper preserves sharper texture boundaries and more faithful colour distributions across objects while remaining within the same or smaller storage budgets.
        \begin{wrapfigure}[4]{L}[0pt]{0.43\linewidth}
            \centering
            \vspace{-12pt}
            \includegraphics[width=\linewidth]{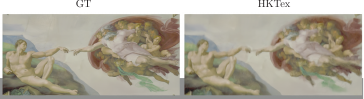}
            \vspace{-22pt}
            \caption{Image on two triangles.}
            \label{fig:image-fit}
        \end{wrapfigure}
        
        \vspace{1mm}\noindent \textbf{Fitting Images.}
            To test \paper's ability to represent high-resolution textures on coarse meshes, we fit an image onto a rectangle made of two triangles.
            In this special case of \uv-texture fitting, we found it beneficial to amplify the weighted and normalised kernels as $(\tilde{h}_t)^a$ before filtering them (with $a=30$). Nevertheless, $a=1$ is sufficient for most meshes.

        \begin{table}[tb]  
            \setlength{\tabcolsep}{4pt}
            \newcommand{\sd}[1]{\,{\scriptstyle \pm #1}} %
            \renewcommand{\cellalign}{bc} %
            \caption{Quantitative comparison of the UV-texture fitting on $313$ meshes from our curated subset of Objaverse~\cite{deitke2023objaverse}. HR and LR indicate high and low resolution.}
            \label{tab:uv-fitting}
            \centering
            \resizebox{\linewidth}{!}{
                \begin{tabular}{@{}l ccccccc @{}}
                \toprule
                & \makecell{PSNR \\ \scriptsize ($\uparrow$)} 
                & \makecell{LPIPS \\ \scriptsize ($\times 10^{-2}, \downarrow$)} 
                & \makecell{SSIM \\ \scriptsize ($\times 10^{-2}, \uparrow$)} 
                & \makecell{MS-SSIM \\ \scriptsize ($\times 10^{-2}, \uparrow$)} 
                & \makecell{MSE \\ \scriptsize ($\times 10^{-3}, \downarrow$)} 
                & \makecell{Render \\ \scriptsize (ms, $\downarrow$)} 
                & \makecell{Storage \\ \scriptsize (KB, $\downarrow$)} \\
                \midrule
                
                LR GT \uv Textures  
                & $\bm{48.4}\sd{\bm{15.2}}$ & $2.3\sd{4.2}$ & $98.2\sd{3.4}$ & $97.7\sd{5.6}$ & $0.89\sd{3.61}$ & \underline{$33.7\sd{3.2}$}& $115.8\sd{103.2}$ \\
                
                GT VTex ($\scal{N} \! \approx \! 7.7k$) 
                & $37.9\sd{8.5}$ & $4.0\sd{5.0}$ & $96.8\sd{4.2}$ & $96.1\sd{6.6}$ & $0.94\sd{2.33}$ & $\bm{25.4} \sd{\bm{3.1}}$& $\bm{17.3}\sd{\bm{32.6}}$ \\
                
                HR GT VTex ($\scal{N} \! \approx \! 210k$) 
                & \underline{$45.3\sd{5.7}$} & $\bm{1.1}\sd{\bm{1.7}}$ & \underline{$98.8\sd{2.1}$} & \underline{$99.4\sd{1.4}$} & $\bm{0.08}\sd{\bm{0.20}}$ & $50.3 \sd{49.4} $& $179.6\sd{89.5}$ \\
                \midrule
                
                MLP Pos. Enc.  
                & $42.9\sd{6.6}$ & $2.3\sd{2.5}$ & $98.2\sd{2.2}$ & $99.0\sd{1.3}$ & \underline{$0.15\sd{0.30}$} & $90.9\sd{35.2}$ & $551.3\sd{759.6}$ \\
                
                InstantNGP \cite{muller2022instant}  
                & $41.3\sd{7.4}$ & $3.2\sd{3.5}$ & $97.5\sd{3.1}$ & $98.2\sd{2.7}$ & $0.26\sd{0.52}$ & $90.8\sd{35.3}$& $473.2\sd{714.9}$ \\
                
                INFs \cite{koestler2022intrinsic}  
                & $42.5\sd{6.8}$ & $2.7\sd{3.4}$ & $98.0\sd{2.7}$ & $98.8\sd{1.9}$ & $0.19\sd{0.44}$ & $91.4\sd{35.3}$ & $928.2\sd{1,439.8}$ \\
                
                ImageGS \cite{zhang2025imagegs}  
                & $44.6\sd{14.6}$ & $2.9\sd{5.4}$ & $97.3\sd{5.6}$ & $95.9\sd{9.6}$ & $3.65\sd{11.95}$ & NA& $115.4\sd{93.9}$ \\
                \midrule 
                
                \paper ($\scal{S} \approx 4.8k$) 
                & $44.8\sd{5.8}$ & \underline{$1.3\sd{1.6}$} & $\bm{98.9}\sd{\bm{1.5}}$ & $\bm{99.5}\sd{\bm{0.7}}$ & $\bm{0.08}\sd{\bm{0.20}}$ & $784.7\sd{474.3}$ & \underline{$96.6\sd{12.2}$} \\
                \bottomrule
                \end{tabular}
            }
        \end{table}

    \subsection{Fitting multi-view images}
        To demonstrate our method's ability to optimize multi-view images, we create a synthetic training scenario in which, instead of UV textures, we use rendered images of the ground-truth UV textures, following the same data setup in UV texture fitting but with a smaller subset of the data. To generate training data, we sample rays from random views during training and select a random subset to form minibatches at each epoch. 
        
        Similar to the UV fitting setup, we implement two MLP architectures conditioned on different positional encodings, corresponding to the spatially varying albedo map. For a fair comparison, these representations are also trained using automatic differentiation and the interoperability between $\mitsuba$ and $\pytorch$. The first baseline utilizes traditional positional encodings. The second baseline is an adaptation of NVDiffRec~\cite{nvdiffrec2022}, with the mesh and environment map fixed to the ground-truth mesh and environment map using hash encoding and with only the albedo map output. Like in the uv-fitting setting, we also conduct a grid search over the width, depth, and the hash-grid size to try to match the storage cost of our model.

        As shown in \cref{tab:ray-fitting}, \paper consistently outperforms the high-resolution vertex colours and neural baselines across all evaluation metrics while requiring substantially less storage. In particular, the improvements in LPIPS and SSIM indicate that our intrinsic formulation preserves perceptual details and structural consistency more effectively during inverse rendering. These results suggest that constraining the representation directly to the surface manifold provides a strong inductive bias for recovering appearance from sparse multi-view observations.
        
        While our representation introduces additional evaluation overhead due to kernel aggregation and geodesic neighbourhood queries as seen in~\Cref{tab:uv-fitting}, this cost is partially amortized in the multi-view setting where ray tracing (with an increased depth) dominates the rendering time as seen in~\Cref{tab:ray-fitting}.

    \begin{table}[tb]  
            \setlength{\tabcolsep}{4pt}
            \newcommand{\sd}[1]{\,{\scriptstyle \pm #1}} %
            \renewcommand{\cellalign}{bc} %
            \caption{Quantitative evaluation of the multi-view rendering on a curated subset of $162$ meshes from Objaverse~\cite{deitke2023objaverse}. NVDiffRec~\cite{nvdiffrec2022} is our adaptation with the ground truth mesh and environment map fixed instead of learned, and with only albedo map output like our method.}
            \label{tab:ray-fitting}
            \centering
            \resizebox{\linewidth}{!}{
                \begin{tabular}{@{}l ccccccc @{}}
                \toprule
                & \makecell{PSNR \\ \scriptsize ($\uparrow$)} 
                & \makecell{LPIPS \\ \scriptsize ($\times 10^{-2}, \downarrow$)} 
                & \makecell{SSIM \\ \scriptsize ($\times 10^{-2}, \uparrow$)} 
                & \makecell{MS-SSIM \\ \scriptsize ($\times 10^{-2}, \uparrow$)} 
                & \makecell{MSE \\ \scriptsize ($\times 10^{-3}, \downarrow$)} 
                & \makecell{Render \\ \scriptsize (s, $\downarrow$)} & \makecell{Storage \\ \scriptsize (KB, $\downarrow$)} \\
                \midrule

                MLP Pos. Enc.  & $ 35.75 \sd{7.76} $ & $3.7 \sd{3.8} $& $97.2 \sd{ 3.1}$& $96.8 \sd{4.9}$& $ 1.55 \sd{4.83} $ & $0.68 \sd{0.12}$ & $ 602.71 \sd{ 755.89} $ \\
                NVDiffRec*~\cite{nvdiffrec2022} & {$36.40 \sd{6.83}$} & {$3.4 \sd{ 3.6}$}& {$97.4 \sd{ 2.9}$}& {$97.4 \sd{ 3.5}$}& {$0.79 \sd{ 2.04}$} & \underline{$0.66 \sd{0.11}$}& {$516.12 \sd{ 722.66}$} \\
                HR VTex
                    & \underline{$37.16 \sd{ 5.69}$} & \underline{$3.1 \sd{ 3.4}$}& \underline{$97.5 \sd{ 2.8}$}& \underline{$97.9 \sd{ 2.6}$}& \underline{$0.44 \sd{ 0.85}$} & $\bm{0.58}\sd{\bm{0.08}}$ & \underline{$81.06 \sd{ 131.16}$} \\
                
                \midrule 
                
                \paper ($\scal{S} \approx 3.8k$) & $\bm{37.61} \sd{\bm{4.71}}$ &  $\bm{2.1} \sd{ \bm{1.9}}$& $\bm{98.3} \sd{ \bm{1.6}}$& $\bm{98.6} \sd{ \bm{1.4}}$& $\bm{0.30} \sd{ \bm{0.53}}$ & $1.2 \sd{0.3}$& $\bm{78.73} \sd{ \bm{23.12}}$ \\
                \bottomrule
                \end{tabular}
            }
        \end{table}

        \EndTightFigureParagraph

    \section{Conclusion}
    \label{sec:conclusion}
    In this work, we presented \paper, an intrinsic texture representation that replaces traditional UV-mapped textures with anisotropic heat kernels defined directly on the surface manifold. By operating entirely in the intrinsic geometry of the mesh, our formulation eliminates seams, distortions, and redundant storage associated with UV parametrisations. Through the use of geodesic heat kernels and Riemannian optimisation, our approach maintains strict alignment with the underlying surface while enabling continuous appearance evaluation within a differentiable physically based renderer. This formulation enables both the compression of existing UV textures and the direct reconstruction of surface appearance from multi-view imagery, while remaining entirely intrinsic to the mesh and independent of any global surface parametrisation.
    HKTex’s mathematical continuity essentially acts as diffusion curves on surfaces, establishing a foundational step toward manifold vector graphics.

    \paragraph{Limitations and future work.}
    Future work may extend the representation beyond albedo to additional appearance attributes such as spatially varying BSDF parameters. While our current evaluation focuses on albedo textures, each kernel could be augmented with additional material parameters (e.g., roughness or specular components) and optimized through differentiable rendering with $\mitsuba$.

    \section*{Acknowledgements}
    T. Birdal was supported by a UKRI Future Leaders Fellowship [grant number MR/Y018818/1].
    S. Foti, S. Zafeiriou, and T. Birdal were supported by the EPSRC Project GNOMON (EP/X011364/1). S. Foti and S. Zafeiriou were also supported by the Turing AI Fellowship MAGAL (EP/Z534699/1).
    
    \bibliographystyle{splncs04}
    \bibliography{main}
    
    \clearpage
\setcounter{page}{1}

\appendix

\renewcommand{\thesubsection}{S.\arabic{subsection}}
\setcounter{subsection}{0} %

\section*{Supplementary Material}

    \subsection{Additional Implementation Details}
        \begin{figure}[h!]
            \centering
            \includegraphics[width=0.8\linewidth]{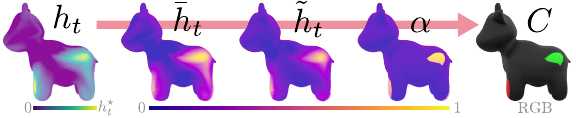}
            \caption{Kernel values at the different stages of the \paper construction process. Where queried points possess only scalar values, we use colourmaps to represent their ranges. For ease of notation, we indicate the source heat as $h_t^\star = h_t\big(\src \!, \src \! \mid \ang, \anis\big)$.}
            \label{fig:process}
        \end{figure}
        
        \paragraph{Visualising heat kernel construction.}
            We here represent the different stages described in \cref{sec:hkt} and comprising the full kernel construction: truncated continuous anisotropic kernels as well as their normalisation, biharmonic distance weighting, filtering, and colour formation. 
            It is interesting to observe the presence of spectral ringing in the early stages; the biharmonic distance weighting successfully dampens these distant artifacts, ensuring their amplitudes fall safely below the cutoff so that the subsequent filtering step ($\alpha$) can completely eliminate them.
        
        \paragraph{Projected Gradient Descent (PGD).}
            To ensure the physical validity and numerical stability of our \paper model, we bound colours, sharpnesses, thresholds, angles, and anisotropies throughout optimisation. Instead of using sigmoid-like activation functions which could potentially lead to vanishing gradients, we optimise the parameters while enforcing bounds using Projected Gradient Descent (PGD). After every optimiser step, we project the parameters back into their valid feasible sets via strict clamping. Specifically, we constrain kernel angles in $\ang \in [0, \pi]$, sharpnesses to $\shr \in [10, 200]$, thresholds to $\thr \in [0.3, 1.0 - 10^{-8}]$, and anisotropies to $ \anis \in [1, 200]$. The bounds for $\shr$, $\thr$, and $\anis$ were empirically selected by observing the kernel behaviours while traversing their values. 
            Additionally, to enable kernels to act as both additive and subtractive high-frequency detail layers over the mean base colour, we allow individual kernel colours to optimise within $\rgb \in [-1, 1]^3$, while the global mean colour is strictly clamped to $\rgb_\text{base}\in[0, 1]^3$. 

        \paragraph{Kernel parameter initialisation.}
            For multi-view image fitting, positions are uniformly sampled over the surface as described in \cref{sec:experiments} and colours are randomly initialised by sampling from a uniform distribution.
            In \uv-texture fitting, $50\%$ of kernel positions are initialised with uniform sampling and $50\%$ according to the \uv-texture content evaluated by computing image gradients like in \cite{zhang2025imagegs} and transferring the sampled 2D points onto the manifold. This $1:1$ ratio of random to image-gradient-based initialisations balances capturing high-frequency details with maintaining sufficient baseline surface coverage. With this strategy, colours are initially sampled from the \uv-texture at the coordinates corresponding to the kernel sources.
            
            Let $u \sim \mathcal{U}(0,1)$ be uniform random variables sampled for each of the $S$ initial kernels.
            For parameters where a uniform distribution is desirable, we simply scale $u$ to the target range. Specifically, kernel angles are initialised as $\ang = u \cdot \pi$, and anisotropies are uniformly sampled in the range $[1, 100]$ via $1.0 + 99.0 \cdot u$. 
            Conversely, we found it beneficial to bias the initialisation of sharpness and threshold parameters towards their higher values. This geometrically biases the kernels to be sharp and highly localised from the very first step, preventing gradient entanglement across overlapping kernels. 
            To achieve this, we sample these parameters from Beta distributions. In practice, we implement this efficiently by applying a fractional power-skew to $u$ before scaling. The kernel sharpnesses $\shr$ are initialised in the range $[10, 50]$ using $\shr = 10 + 40 \cdot u^{0.5}$. Similarly, the evaluation thresholds $\thr$ are initialised in the range $[\thr_{\text{min}}, 1.0)$ using $\thr = \thr_{\text{min}} + (1.0 - \thr_{\text{min}} - 10^{-8}) \cdot u^{0.7}$, where $\thr_{\text{min}}$ is a configurable lower bound (set to $0.7$ and $0.9$ for our \uv-texture and multi-view fitting experiments, respectively).

        \paragraph{Importance sampling.}
            To ensure we prioritise the optimisation of regions of the surface that are harder to optimise for their richness of details, we adopt an importance sampling strategy. Before initiating the optimisation procedure, we uniformly pre-sample a pool of $10M$ points on the surface of the mesh. These points are used throughout the optimisation to evaluate $C(\pt)$ as in \cref{eq:color_formation} and assess the photometric loss $\mathcal{L}(\pt)$ against the GT colour $C_\text{GT}(\pt)$. During the first $250$ optimisation steps, samples are uniformly drawn from the pool. After this warm-up phase, the errors acting as sampling weight $\varepsilon(\pt)$ associated with each point starts to be recorded using an exponential moving average update $\varepsilon_\text{new}(\pt) = \gamma \varepsilon_\text{old}(\pt) + (1 - \gamma) \mathcal{L}(\pt)$ with decay $\gamma=0.99$. These $\varepsilon(\pt)$ are thus used to drive the sampling distribution towards points with higher reconstruction errors.
            
        \paragraph{Mass kernel density estimation.}
            As mentioned in \cref{sec:spatial_eigen_and_masses}, the evaluation of \cref{eq:cont_heat_kernel} requires the mass $m^\star$ at the continuous source point $\src$. While in our application, setting it to a unit value does not affect performance, we formulate a more general Kernel Density Estimation (KDE) approach for the mass estimation at arbitrary locations on the surface of a mesh. 
            The intuition for the KDE mass is that points that cluster closely together should represent a smaller fraction of the surface area, while isolated points should represent larger areas. We thus start by evaluating the pairwise biharmonic distances~\cite{lipman2010biharmonic} between all sources. Then, we apply a Gaussian kernel to the pairwise distance matrix $\mat{D}$ and sum across rows to compute the local density ($\varrho_s$) for each source:
            \begin{equation*}
                \varrho_s = \sum_{i=1}^\scal{S} \exp \Big(- \frac{\scal{D}_{i,s}^2}{\sigma_m^2}\Big),
            \end{equation*}
            with $\sigma_m^2$ being an adaptive bandwidth computed by taking the median of all non-zero distances.
            Masses are thus defined as the inverse of the density: $m^\star_s = \frac{1}{\varrho_s}$. Finally, the masses undergo an additional normalisation step enforcing their sum to equal the total area of the surface. 
            
            This estimate can also be used for diffusing heat to points that differ from the sources. In this case, arbitrary points can be treated as additional sources without altering the estimation procedure. 

        \paragraph{Visualising heat kernels.}
            In order to visualise the distribution of heat kernels over the surface of each mesh, we modify the filtering function introduced in \cref{eq:hk_filtering} and replace it with a boxcar function. This simple modification allows us to trace hollow geodesic ellipsoids. To ease visualisation, we also randomise their colour. Although in most cases a high density of kernels is leveraged to fit subtle albedo variations, when uniform colours are present, \paper can adaptively adjust the position and size of kernels by increasing sparsity in such regions (see the Earth example in \cref{fig:qual results}).  
            
        \paragraph{Additional details on kernel splitting.}
            Since the anisotropic heat kernels of
            {\parfillskip0pt\par}
            \begin{wrapfigure}[9]{L}[0pt]{0.52\linewidth}
                \centering
                \vspace{-25pt}
                \includegraphics[width=\linewidth]{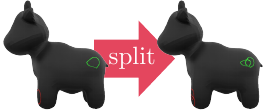}
                \caption{Visualisation of kernel split performed on two kernels. }
            \end{wrapfigure}
            \noindent
            \paper are defined with respect to the principal curvatures (\cref{sec:background}), it is natural to use the same axes also during the kernel densification strategies requiring a kernel split. Like many other quantities encountered thus far, the principal curvature of a surface discretised as a mesh is also defined only at vertex locations $\mat{V}_M \in \R^{N \times 3}= [\versor[v]{M}^{(1)},\dots,\versor[v]{M}^{(N)}]^\top$. Therefore, the first step requires the barycentric interpolation of the principal direction at the old source position: $\versor[v]{M}^\star=b^{\src_\text{old}} (\mat{V}_M)$. Since the kernel $\hk_\text{old}$ centred at $\src_\text{old}$ is rotated according to its learnable $\ang_\text{old}$, we define a new major axis vector rotating $\versor[v]{M}^\star$ by $\ang_\text{old} + \pi/2$ around the face normal ($\versor[n]{}$), obtaining:           
            \begin{equation*}
                \versor[d]{}^\star = \versor[v]{M}^\star \cos\Big(\ang_\text{old} + \frac{\pi}{2}\Big) + (\versor[n]{} \times \versor[v]{M}^\star) \sin\Big(\ang_\text{old} + \frac{\pi}{2}\Big).
            \end{equation*}
            The old kernel is then removed and replaced by two new kernels displaced according to the exponential map computed with~\cite{verninas2026disgeod} in opposite directions along this major kernel axis. Specifically, we compute:
            \begin{equation*}
                \pt^{\star (\pm)}_{\text{new}} = \Exp \Big(\src_\text{old}, \pm \, \versor[d]{}^\star (1 - \thr_\text{old}) r_\text{base}\Big),
            \end{equation*}
            where $r_\text{base}$ is a base scaling radius, $(1 - \thr_\text{old})$ acts as a proportional proxy for the kernel's spatial radius, and the sign used for computing the exponential map ($\pm$) determines the two new locations.
            
            Finally, the newly split kernels must be shrunk. Their spatial extents are reduced by a factor of $1.6$, yielding new thresholds $\thr_\text{new} = 1.0 - \frac{1.0 - \thr_\text{old}}{1.6}$. Furthermore, the optimiser states (e.g., momentum and variance) for these new parameters are strictly zeroed out to prevent the injected kernels from inheriting the extreme gradients that triggered the split.

        \paragraph{Additional details on rendering and multi-view optimisation.}
            For multi-view optimisation, we instantiate \paper as a custom differentiable $\mitsuba$ texture~\cite{jakob2022mitsuba3} and attach it to a Principled BSDF~\cite{Disney2012, Disney2015} on the target mesh, currently as the albedo texture. 
            During training, we use a perspective camera with a sensor capturing images of $256\times256$ pixels, a constant emitter, and a path-replay-backpropagation (PRB)-style ray integrator (\texttt{rb\_ray}) built on top of Mitsuba's PRB integrator~\cite{Vicini2021PathReplay}. 
            Our implementation of this custom PRB integrator is based on the existing extension of~\cite{wiersma2025uncertainty}.
            
            The multi-view optimisation is organised as a two-level loop: an outer iteration samples a fresh set of camera views and constructs the pool of intersecting rays, while the corresponding inner iterations process that pool through successive ray mini-batches and apply the parameter updates.
            At each outer training iteration, we sample a batch of random cameras by uniformly drawing azimuth, elevation, and camera distance within the configured bounds. 
            For each sampled camera, we then generate one primary ray per pixel and iteratively resample until retaining only rays that actually intersect the object. 
            The resulting ray set from all views is concatenated, randomly shuffled, and partitioned into mini-batches of $1024$ rays. 
            This corresponds to a total of $3840$ optimisation steps in the reported multi-view setting, counting all inner ray-mini-batch updates rather than only the outer camera-sampling iterations.

            During optimisation, each selected ray is re-evaluated with configurable gradient samples per pixel ($N_{\text{spp}}=8$ are used in our results). When performing synthetic training on models with an existing UV texture, the ground-truth images are rendered with the same integrator and then integrated across the same samples. 
            Before computing the photometric loss, both prediction and target are clipped to $[0,1]$ and converted from linear radiance to gamma space, matching the image-space supervision used in our experiments. During this step, clipping can be omitted if the target images use High Dynamic Range (HDR).
            In the reported multi-view setting, we optimise a per-ray photometric loss corresponding to the L1 loss averaged across the ray mini-batch. 
            Formally, let $\mathcal{R}=\{r_i\}_{i=1}^{B}$ be the current mini-batch of intersecting rays, with $B=1024$. For each ray $r_i$, the renderer produces $N_{\text{spp}}$ Monte Carlo radiance samples $\{\mathbf{L}_{i,s}\}_{s=1}^{N_{\text{spp}}}$, which are averaged to obtain the predicted linear RGB colour
            \begin{equation*}
                \hat{\mathbf{c}}_i = \frac{1}{N_{\text{spp}}}\sum_{s=1}^{N_{\text{spp}}} \mathbf{L}_{i,s}.
            \end{equation*}
            Denoting by $\mathbf{c}'_i$ the corresponding target colour, we apply the same image-space preprocessing to both prediction and target:
            \begin{equation*}
                \tilde{\mathbf{c}}_i = \Gamma\big(\operatorname{clip}(\hat{\mathbf{c}}_i,0,1)\big),
                \qquad
                \tilde{\mathbf{c}}'_i = \Gamma\big(\operatorname{clip}(\mathbf{c}'_i,0,1)\big),
            \end{equation*}
            where $\operatorname{clip}(\cdot,0,1)$ is applied channel-wise and $\Gamma(\cdot)$ denotes a linear-to-gamma colour transformation. The ray-training loss is then:
            \begin{equation*}
                \Loss_{\text{ray}} = \frac{1}{B} \sum_{i=1}^{B} \left\| \tilde{\mathbf{c}}_i - \tilde{\mathbf{c}}'_i \right\|_1,
            \end{equation*}
            which corresponds to the L1 configuration used in our multi-view experiments.
            For efficiency, the heat-kernel texture queries are evaluated in chunks (typically $1024$ surface points at a time), and for final renderings, we optionally tile the image plane (e.g., $32\times32$ crops) to limit GPU memory.
            We use the OptiX denoiser~\cite{optix, optixdenoiser} only for the displayed renderings, and not in the ray-based training loss itself. 

            Note that although we train using synthetic rendered images, the ground-truth target signal could naturally be replaced with screen-space image pixels from real photographs to perform multi-view image training. However, because our current implementation parametrises only the albedo component of the Principled BSDF, optimising directly against real photographs would cause the model to incorrectly ``bake'' scene lighting, shadows, and view-dependent reflections into the diffuse colour. Fully leveraging a Physically Based Rendering (PBR) pipeline requires disentangling these effects. While extending our \paper formulation to jointly optimise additional material properties (e.g., roughness, metallic, and normal maps) is highly feasible, we currently isolate the albedo fitting using synthetic ground truth to ensure physically accurate material reconstruction.

        \paragraph{Hyperparameters and hyperparameter tuning.}
            As mentioned in \cref{sec:experiments}, to identify the best hyperparameters for both our \uv~and multi-view experiments, we performed a Bayesian hyperparameter optimisation using $\optuna$~\cite{akiba2019optuna} on $5$ meshes from our curated subset of the Objaverse~\cite{deitke2023objaverse}. These meshes were excluded from all evaluation sets. 

            Our hyperparameter search was formulated as a multi-objective optimisation problem, with the Tree-Structured Parzen Estimator~\cite{bergstra2011tpe} minimising both the final model storage size and the mean squared errors across $3$ rendered views of the tuning meshes.  Because the optimisation dynamics differ between our tasks, we executed separate tuning sweeps for \uv-texture fitting and multi-view image fitting. 
            
            The search space encompassed over 30 parameters. Continuous variables, such as learning rates for the kernel parameters and the density controller thresholds, were sampled from log-uniform distributions. Discrete categorical variables were used to select optimal architectural choices, including initialisation strategies, loss functions (e.g., MSE, Smooth L1, L1), and learning rate schedulers (Step, Cosine, or none). The final list of hyperparameters is reported in \cref{tab:hyperparameters}.

            \begin{table}
                \centering
                \caption{\textbf{Hyperparameter configurations.} We report the exact hyperparameters used for our \uv-texture fitting and multi-view image fitting experiments. All parameters except the Optimisation Steps were dynamically selected from the search space via our multi-objective Ray Tune and Optuna optimisation.}
                \label{tab:hyperparameters}
                \resizebox{\textwidth}{!}{
                \begin{tabular}{llccc}
                    \toprule
                    \textbf{Category} & \textbf{Parameter} & \textbf{UV} & \textbf{Multi-View} \\
                    \midrule
                    \multirow{4}{*}{General} 
                    & Optimisation Steps & 5,000 & 3,840 \\
                    & Batch Size & 4,096 (points) & 1,024 (rays) \\
                    & Loss Function & MSE & L1 \\
                    & Random:Gradient Init. & 1:1 & NA \\
                    \midrule
                    \multirow{5}{*}{Architecture} 
                    & Initial Sources ($S$) & 1,000 & 5,000 \\
                    & Diffusion Time ($t$) & $6.25 \times 10^{-2}$ & $1.11 \times 10^{-6}$ \\
                    & Colour Formation k-NN ($\scal{K}_c$) & 30 & 20 \\
                    & Sources k-NN ($\scal{K}_s$) & 50 & 50 \\
                    & Init. Threshold ($\thr_\text{min}$) & 0.3 & 0.7  \\
                    \midrule
                    \multirow{10}{*}{Learning Rates} 
                    & Mean Colour ($\rgb_\text{base}$) & $9.61 \times 10^{-4}$ & $1.71 \times 10^{-3}$  \\
                    & Kernel Colours ($\rgb$) & $6.15 \times 10^{-3}$ & $3.33 \times 10^{-2}$  \\
                    & Angles ($\ang$) & $9.93 \times 10^{-2}$ & $1.07 \times 10^{-4}$ \\
                    & Anisotropies ($\anis$) & $4.67 \times 10^{-3}$ & $3.66 \times 10^{-2}$ \\
                    & Thresholds ($\thr$) & $1.09 \times 10^{-3}$ & $1.36 \times 10^{-3}$ \\
                    & Sharpnesses ($\shr$) & $1.29 \times 10^{-3}$ & $1.20 \times 10^{-3}$  \\
                    & Locations ($\src$) & $9.91 \times 10^{-2}$ & $6.30 \times 10^{-2}$  \\
                    & Geodesic Momentum & 0.9 & 0.0 \\
                    & Adam Scheduler & StepLR & StepLR  \\
                    & Geodesic Scheduler & StepLR & Cosine Annealing \\
                    \midrule
                    \multirow{5}{*}{Density Control} 
                    & Max Kernels Limit & 5,000 & 10,000 \\
                    & Pruning Selection Threshold & $1.69 \times 10^{-3}$ & $2.49 \times 10^{-3}$ \\
                    & Pruning Contrib. Threshold & $3.42 \times 10^{-2}$ & $2.05 \times 10^{-2}$ \\
                    & Densify Error Threshold & $2.37 \times 10^{-2}$ & $2.38 \times 10^{-2}$  \\
                    & Densify Size Threshold & 0.174 & 0.181 \\
                    \bottomrule
                \end{tabular}}
            \end{table}

        \paragraph{A note on the eigenproperties precomputation.}
            In \cref{sec:precomp_interp_eigen} we detailed the precomputation of the ALBO eigenproperties ($\Evecs_{i,j}^\ddag$ and $\Evals_{i,j}^\ddag$) on the half-polar-like grid. While the storage cost of these eigenproperties across all the $7\times7$ nodes in our grid averages $\sim 370$MB per mesh, they are a deterministically computed quantity and they do not need to be stored. 
            We precompute the eigendecompositions and cache them once per mesh and reuse them across runs, and don't include the precomputations in the timing. 
            Precomputing the ALBO eigendecompositions for the $7 \times 7$ half-polar grid takes $(97.8 \pm 286.9)$s. Of this, the Hungarian method required only $(0.05 \pm 0.03)s$ to align the $49$ eigenvectors $\Evecs_{i,j}^\ddag$. This (deterministic) sequential CPU overhead could be significantly optimised via GPU parallelization, making on-device recomputation fast enough to completely eliminate the need for transmission while preserving our storage benefits. We leave this for a future study.

        \StartTightFigureParagraph
        \paragraph{A note on the eigenmodes and vertex count dependence.}
            \paper is influenced by the number of eigenmodes, but $K=256$ empirically offered sufficient 
            {\parfillskip0pt\par}
            \begin{wrapfigure}[7]{R}[0pt]{0.24\linewidth}
                \centering
                \vspace{-10pt}
                \includegraphics[width=\linewidth]{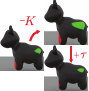}
            \end{wrapfigure}
            \noindent            
            control in our experiments. Kernel size increases when reducing $K$, but the learnable $\thr$ can compensate (see insert).
            
            Pushing \paper to the extreme while fitting an image on a rectangle made of two triangles (\cref{fig:image-fit}), and thus having $K=3$ proves that our method is not constrained by the vertex count and can represent textures even at higher resolutions than the mesh. Meshes in our test data have $\scal{N} \approx (9.6\pm19)k$ vertices, but we outperform alternative methods (\cref{tab:uv-fitting}) with just $\scal{S} \approx (4.3 \pm 0.6)k$ kernels. $49.8\%$ of the faces on average have no kernels, while $13.2\%$ have one, but the average number of kernel centres per face is $4.8 \pm 7.1$, and the maximum reaches $70 \pm 123$. This proves how our kernels can move where they benefit texture quality the most. 
        \EndTightFigureParagraph

        \paragraph{A note on the \paper editability.}
            While direct editing is non-trivial (a limitation shared with GS, where Spherical Harmonics further exacerbate the issue), modifications can be achieved by baking the target region to a dense point cloud, painting it, and running a brief optimization to update the \paper parameters.

    \subsection{Additional Experiments}
\begin{figure}
                \centering
                \includegraphics[width=1\linewidth]{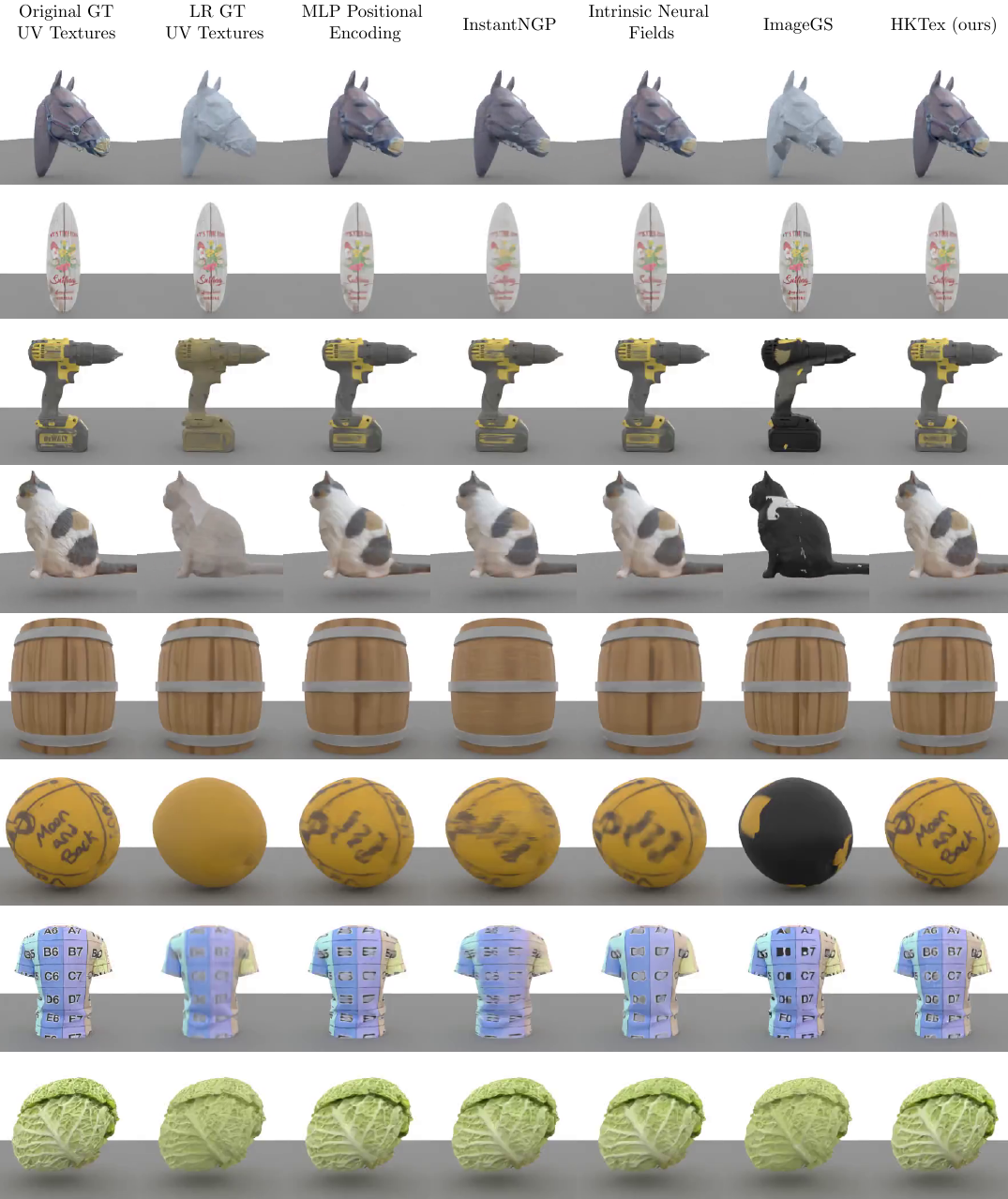}
                \caption{Additional qualitative comparison against \sota~ methods when fitting \uv-textures. With the exception of the GT \uv Textures, all other methods have similar or bigger storage size.
                }
                \label{fig:qual_comparison_suppl1}
            \end{figure}

\begin{figure}
                \centering
                \includegraphics[width=1\linewidth]{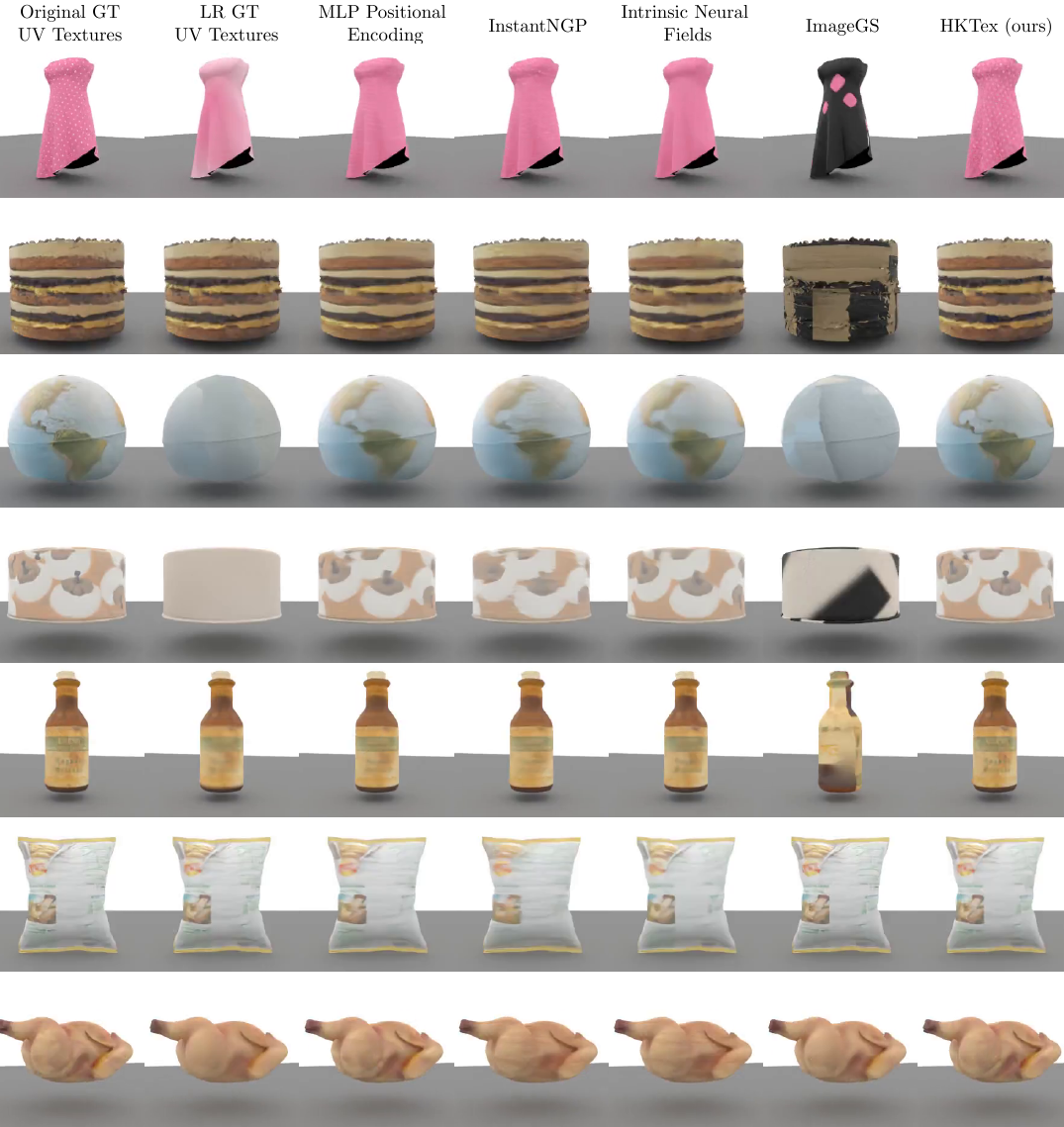}
                \caption{Additional qualitative comparison against \sota~ methods when fitting \uv-textures. With the exception of the GT \uv Textures, all other methods have similar or bigger storage size. 
                }
                \label{fig:qual_comparison_suppl2}
            \end{figure}

\begin{figure}
                \centering
                \includegraphics[width=1\linewidth]{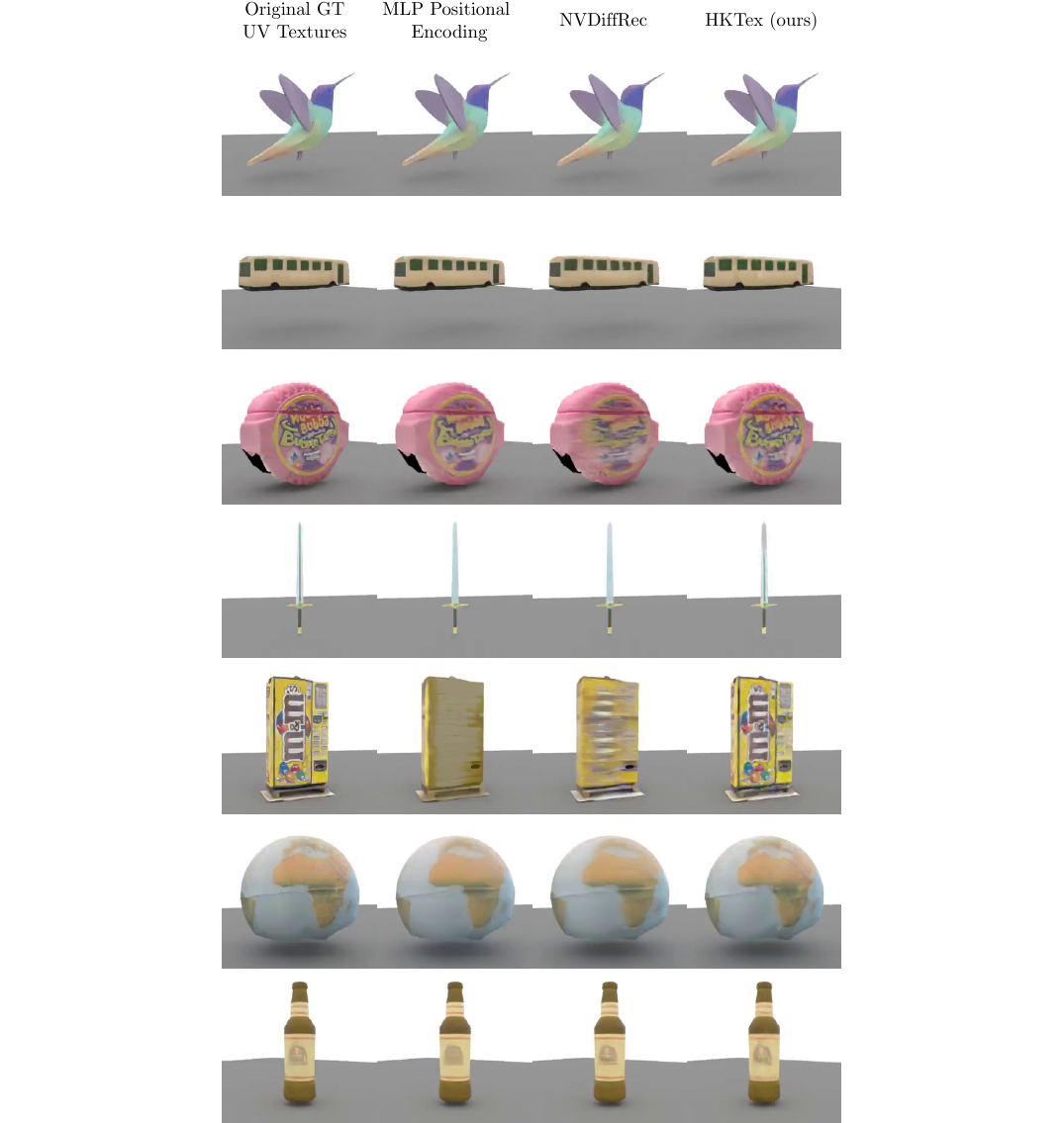}
                \caption{Qualitative comparison against \sota~ methods when fitting synthetic multi-view images. With the exception of the GT \uv Textures, all other methods have similar or bigger storage size. 
                }
                \label{fig:qual_comparison_mv}
            \end{figure}

\begin{figure}
                \centering
                \includegraphics[width=1\linewidth]{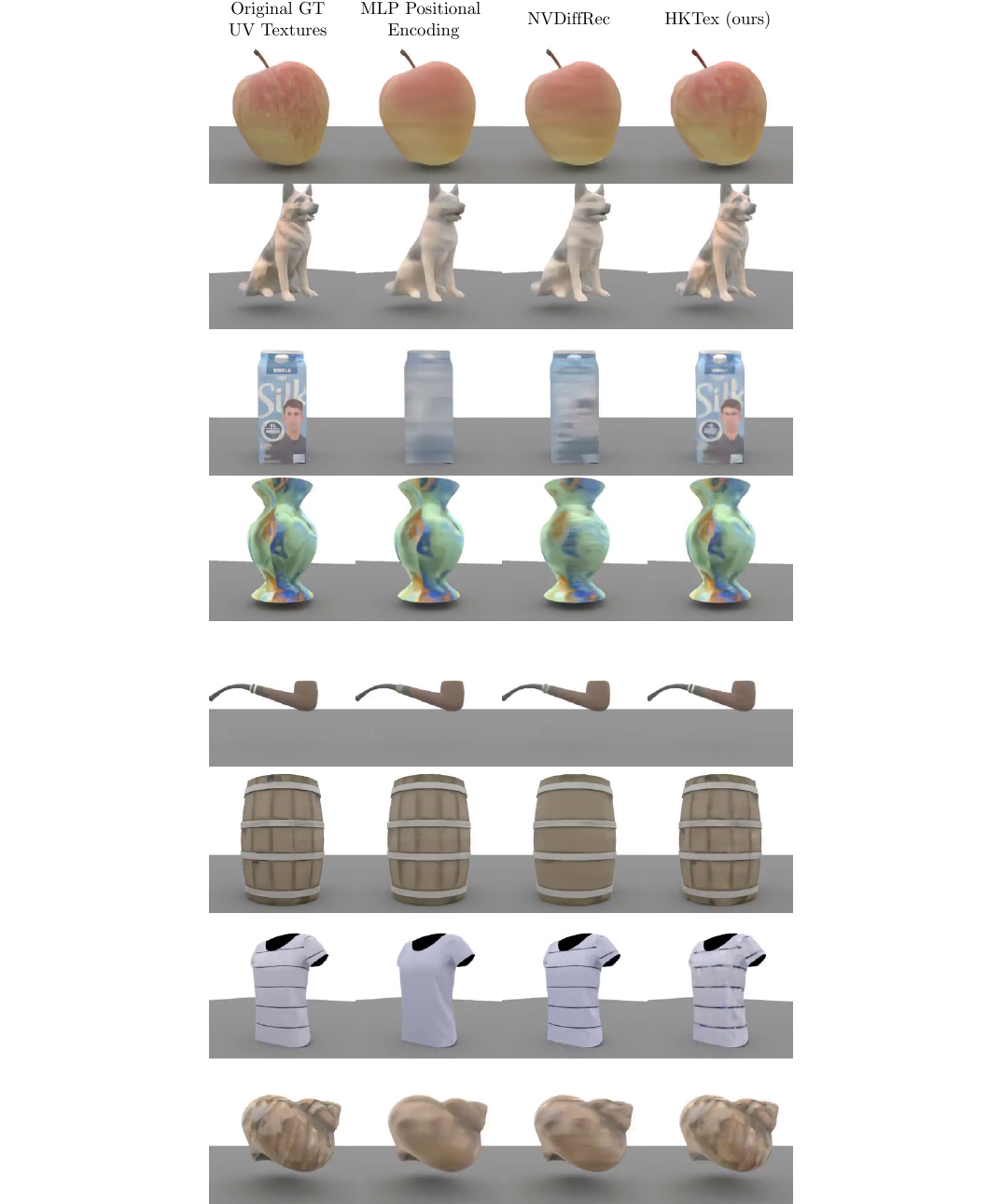}
                \caption{Additional qualitative comparison against \sota~ methods when fitting synthetic multi-view images. With the exception of the GT \uv Textures, all other methods have similar or bigger storage size. 
                }
                \label{fig:qual_comparison_mv2}
            \end{figure}

        In this section, we perform a comprehensive ablation of the main components comprising our method.
        In addition, we report numerous qualitative results obtained by fitting our method to \uv textures (\cref{fig:qual results}) as well as additional comparisons against \sota~\\methods on both the \uv-texture (\cref{fig:qual_comparison_suppl1,fig:qual_comparison_suppl2}) and multi-view image fitting settings (\cref{fig:qual_comparison_mv,fig:qual_comparison_mv2}). Our method significantly outperforms other methods by producing sharper and more detailed results that are not affected by \uv-coordinates overheads. In general, \paper closely matches the visual quality of the original GT textures.           
        
        \begin{table}[b]
            \centering
            \caption{Ablation results on a smaller subset of Objaverse~\cite{deitke2023objaverse} on UV texture fitting. Best values are in bold and second-best values are underlined. Arrows indicate whether lower or higher is better.}
            \small
            \label{tab:ablation}
            \resizebox{\linewidth}{!}{\begin{tabular}{@{}lccccccc@{}}
                \toprule
                Model & PSNR ($\uparrow$) & LPIPS ($\downarrow$) & SSIM ($\uparrow$) & MS-SSIM ($\uparrow$) & MSE ($\times 10^{-3} \!,\downarrow$) & Storage (KB, $\downarrow$) & Kernels ($\downarrow$) \\
                \midrule
                Full Model
                & 45.05 $\pm$ 4.90
                & \textbf{0.011 $\pm$ 0.010}
                & \textbf{0.991 $\pm$ 0.008}
                & \textbf{0.996 $\pm$ 0.004}
                & \textbf{0.058 $\pm$ 0.082}
                & 96.87 $\pm$ 11.90
                & 4811 $\pm$ 557 \\
                \midrule
                No Error-Based Densification
                & 43.05 $\pm$ 5.61
                & 0.020 $\pm$ 0.019
                & 0.985 $\pm$ 0.015
                & 0.992 $\pm$ 0.008
                & 0.104 $\pm$ 0.146
                & \textbf{22.18 $\pm$ 1.01}
                & \textbf{995 $\pm$ 42} \\
                 No Pruning
                & \underline{45.06 $\pm$ 4.91}
                & \textbf{0.011 $\pm$ 0.010}
                & \textbf{0.991 $\pm$ 0.008}
                & \textbf{0.996 $\pm$ 0.004}
                & \textbf{0.058 $\pm$ 0.083}
                & 96.89 $\pm$ 11.94
                & 4810 $\pm$ 558 \\
                No Density Control
                & 43.09 $\pm$ 5.60
                & 0.020 $\pm$ 0.019
                & 0.985 $\pm$ 0.015
                & 0.992 $\pm$ 0.009
                & 0.102 $\pm$ 0.142
                & \underline{22.29 $\pm$ 0.54}
                & \underline{1000 $\pm$ 0} \\
                No Colour Formation KNN
                & 45.04 $\pm$ 4.92
                & \textbf{0.011 $\pm$ 0.010}
                & \textbf{0.991 $\pm$ 0.008}
                & \textbf{0.996 $\pm$ 0.004}
                & \textbf{0.058 $\pm$ 0.080}
                & 96.75 $\pm$ 11.98
                & 4804 $\pm$ 561 \\
                No Biharmonic Weighting
                & 43.99 $\pm$ 5.71
                & 0.016 $\pm$ 0.015
                & 0.988 $\pm$ 0.012
                & 0.994 $\pm$ 0.006
                & 0.093 $\pm$ 0.154
                & 95.59 $\pm$ 9.48
                & 4896 $\pm$ 427 \\
                No Importance Sampling
                & \textbf{45.13 $\pm$ 5.86}
                & \underline{0.012 $\pm$ 0.022}
                & 0.988 $\pm$ 0.033
                & 0.993 $\pm$ 0.027
                & 0.691 $\pm$ 6.148
                & 94.39 $\pm$ 18.38
                & 4712 $\pm$ 836 \\
                No PGD
                & 44.22 $\pm$ 5.10
                & 0.014 $\pm$ 0.013
                & \underline{0.989 $\pm$ 0.010}
                & \underline{0.995 $\pm$ 0.005}
                & \underline{0.071 $\pm$ 0.091}
                & 96.93 $\pm$ 12.84
                & 4853 $\pm$ 517 \\
                \bottomrule
            \end{tabular}}
        \end{table}

        \paragraph{Ablation}
            In~\Cref{tab:ablation}, we provide ablation results obtained by removing different components of our \paper on the UV-texture fitting setting, evaluated on a subset of 92 meshes from Objaverse~\cite{deitke2023objaverse}, filtered as previously described. The rows in \Cref{tab:ablation} correspond to removing one component at a time from the full model: \emph{No Error-Based Densification} disables the insertion of new kernels from the accumulated reconstruction error, \emph{No Pruning} disables the pruning step, \emph{No Density Control} removes both densification and pruning, \emph{No Colour Formation KNN} disables the colour-formation-KNN stage, \emph{No Biharmonic Weighting} disables the distance-based attenuation used to enforce local support, \emph{No Importance Sampling} replaces our importance sampling with uniform sampling, and \emph{No PGD} removes the projected range constraints on the optimised parameters. 
            In this non-PGD variant, bounded geometric parameters are instead enforced with sigmoid activations, while kernel colours use a \texttt{tanh} activation.

            Overall, the tuned full model is primarily driven by adaptive kernel insertion rather than pruning alone. We observe that removing pruning leaves the final quality almost unchanged, indicating that error-based densification together with manifold-aware kernel splitting already provides sufficient adaptivity. 
            By contrast, removing error-based densification causes an immediate degradation, and removing density control altogether does not recover this loss, confirming that adaptive kernel insertion is the dominant density-control mechanism. 
            Disabling the colour-formation-KNN stage yields results that are effectively tied to those of the full model, suggesting that this component is not critical for accuracy. However, it can still offer small computational benefits and is therefore left as an implementation choice. 
         
            The remaining ablations are also consistent with the intended roles of the corresponding components. Removing the biharmonic weighting degrades all image metrics, as expected from its role in suppressing long-range spectral ringing and enforcing local support. Removing importance sampling yields competitive PSNR but much worse MSE, suggesting that uniform sampling overfits easier regions while failing to consistently resolve harder, high-frequency areas. Removing PGD also lowers performance, supporting our decision to constrain kernel parameters throughout optimisation explicitly.

            \begin{figure}
                \centering
                \includegraphics[width=1\linewidth]{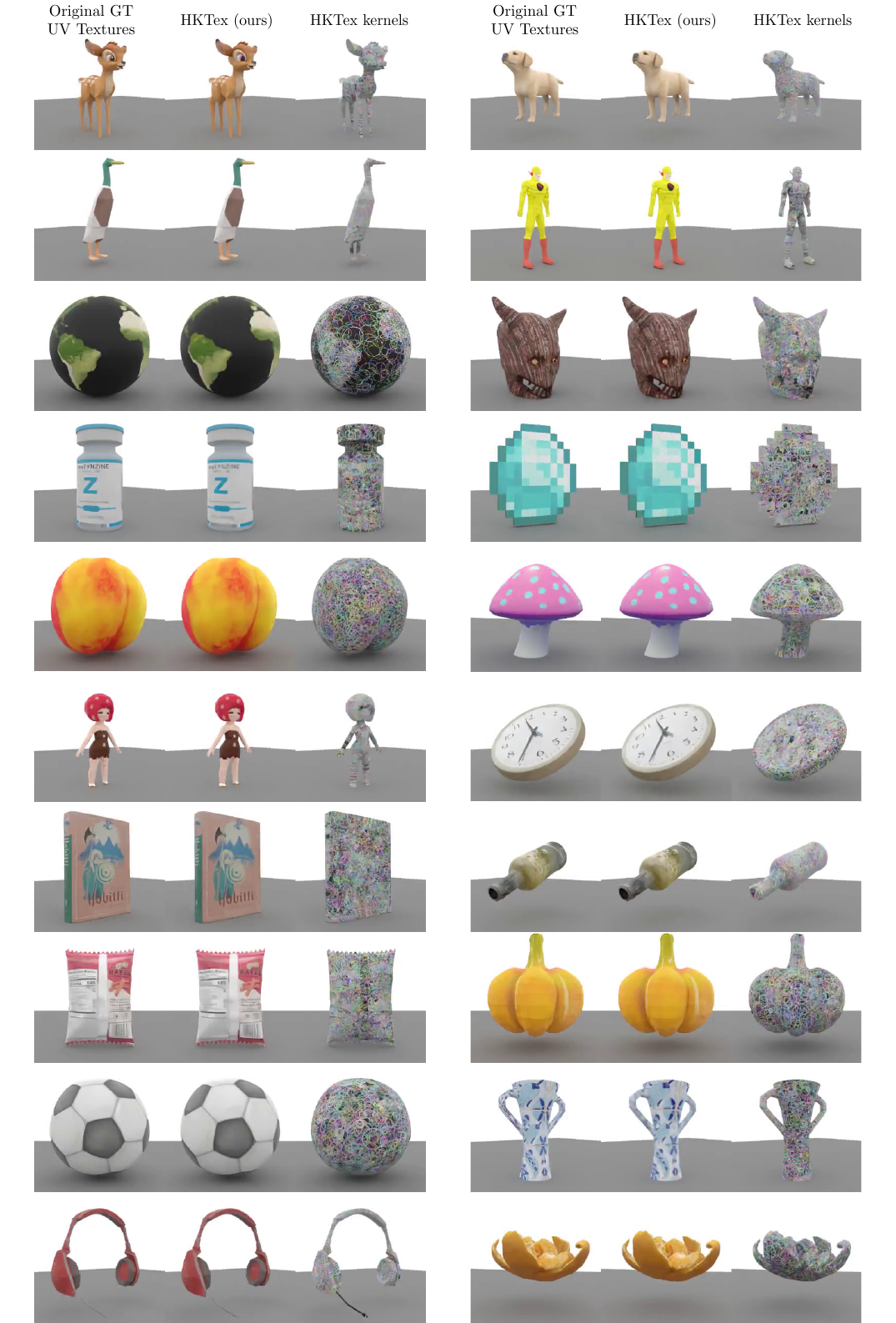}
                \caption{Qualitative of our \paper when fitting \uv-textures. Our textures require approximately one order of magnitude less storage than the original GT textures.
                }
                \label{fig:qual_res}
            \end{figure}
            
\end{document}